\def\year{2021}\relax
\documentclass[letterpaper]{article} 
\usepackage{aaai21}  
\usepackage{times}  
\usepackage{helvet} 
\usepackage{courier}  
\usepackage[hyphens]{url}  
\usepackage{graphicx} 
\usepackage{natbib}  
\usepackage{caption} 
\usepackage{xcolor}
\usepackage{kotex}
\usepackage{blindtext}

\usepackage{gensymb}
\usepackage{amsmath}
\usepackage{amssymb}
\usepackage{amsthm}
\usepackage{exscale}
\usepackage{textcomp}		
\usepackage{enumitem}

\usepackage{array}
\usepackage{tabulary}
\usepackage{multirow}
\usepackage{ctable} 
\usepackage{booktabs}	
\usepackage[toc,title,page]{appendix}

\usepackage[switch]{lineno}

\usepackage[flushleft]{threeparttable}

\newcommand\qa{17,983}
\newcommand\trainqa{11,118}
\newcommand\valqa{3,412}
\newcommand\testqa{3,453}
\newcommand\vidhr{20.5}
\newcommand\visual{217,308}
\newcommand\clips{23,928}
\newcommand\modelname{Multi-level Context Matching}

\title{DramaQA: Character-Centered Video Story Understanding with Hierarchical QA}
\author {
    Seongho Choi,\textsuperscript{\rm 1}
    Kyoung-Woon On,\textsuperscript{\rm 1}
    Yu-Jung Heo,\textsuperscript{\rm 1} \\
    Ahjeong Seo,\textsuperscript{\rm 1}
    Youwon Jang,\textsuperscript{\rm 1}
    Minsu Lee,\textsuperscript{\rm 1}
    Byoung-Tak Zhang\textsuperscript{\rm 1,2}\\
}
\affiliations {
    \textsuperscript{\rm 1} Seoul National University 
    \textsuperscript{\rm 2} AI Institute (AIIS)\\
    \{shchoi,kwon,yjheo,ajseo,ywjang,mslee,btzhang\}@bi.snu.ac.kr
}

\begin{document}

\maketitle

\begin{abstract}

Despite recent progress on computer vision and natural language processing, developing a machine that can understand video story is still hard to achieve due to the intrinsic difficulty of video story. Moreover, researches on how to evaluate the degree of video understanding based on human cognitive process have not progressed as yet. 
In this paper, we propose a novel video question answering (Video QA) task, DramaQA, for a comprehensive understanding of the video story. 
The DramaQA focuses on two perspectives:
1) Hierarchical QAs as an evaluation metric based on the cognitive developmental stages of human intelligence. 2) Character-centered video annotations to model local coherence of the story. 
Our dataset is built upon the TV drama ``Another Miss Oh"\footnote{We have received an official permission to use these episodes for research purposes from the content provider.} and it contains {\qa} QA pairs from {\clips} various length video clips, with each QA pair belonging to one of four difficulty levels.
We provide {\visual} annotated images with rich character-centered annotations, including visual bounding boxes, behaviors and emotions of main characters, and coreference resolved scripts. Additionally, we suggest {\modelname} model which hierarchically understands character-centered representations of video to answer questions. 
We release our dataset and model publicly for research purposes\footnote{https://dramaqa.snu.ac.kr}, and we expect our work to provide a new perspective on video story understanding research.
\end{abstract}
 
\begin{figure*}[h]
 \centering
 \includegraphics[width=1.0\textwidth]{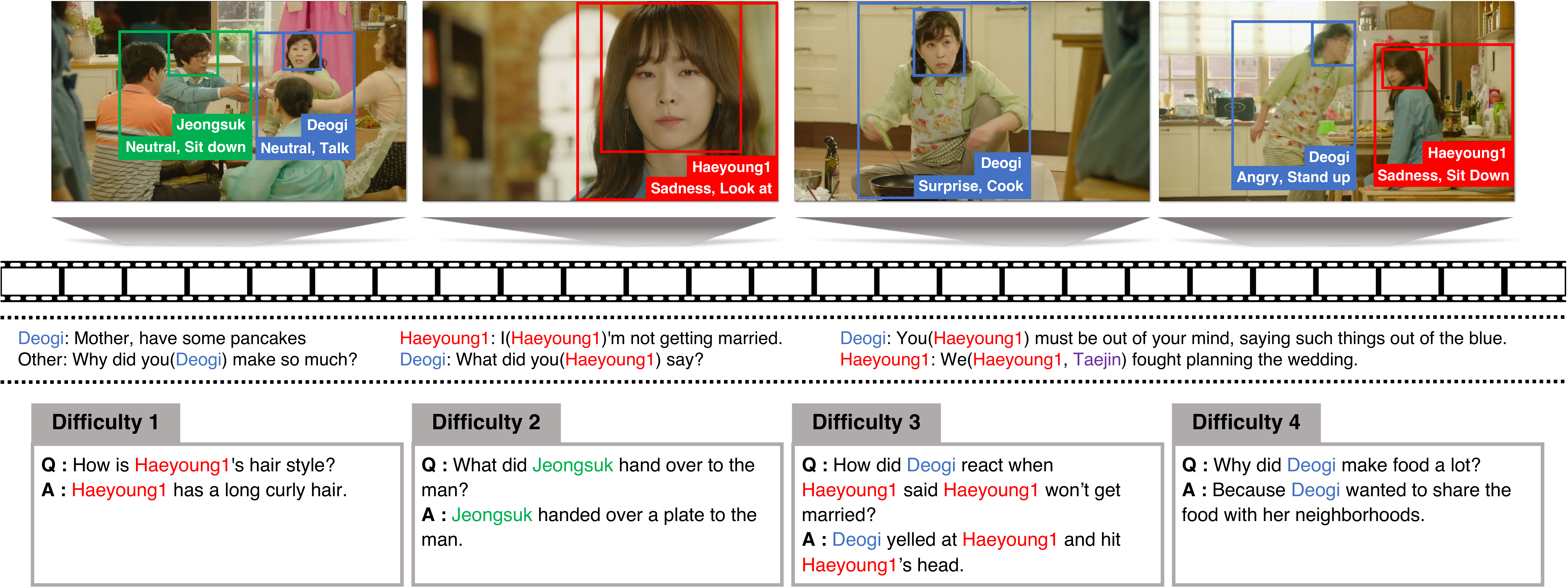}
 \caption{An example of DramaQA dataset which contains video clips, scripts, and QA pairs with levels of difficulty. A pair of QA corresponds to either a shot or a scene, and each QA is assigned one out of possible four stages of difficulty (details in Section DramaQA Dataset). A video clip consists of a sequence of images with visual annotations centering the main characters.}
 \label{fig:dramaqa_overview}
 \end{figure*}
 \index{figure}
 
\section{Introduction}

Stories have existed for a long time with the history of mankind, and always fascinated humans with enriching multimodal effects from novels to cartoons, plays, and films.
The story understanding ability is a crucial part of human intelligence that sets humans apart from others~\cite{szilas1999interactive,winston2011strong}.

To take a step towards human-level AI, \textit{drama}, typically in the form of video, is considered as proper mediums because it is one of the best ways to covey a story. Also, the components of drama including image shots, dialogues, sound effects, and textual information can be used to build artificial ability to see, listen, talk, and respond like humans. 
Since drama closesly describes our everyday life, the contents of drama also help to learn realistic models and patterns of humans’ behaviors and conversations.  
However, the causal and temporal relationships between events in drama are usually complex and often implicit~\cite{riedl2016computational}. 
Moreover, the multimodal characteristics of the video make the problem trickier. 
Therefore, video story understanding has been considered as a challenging machine learning task.

One way to enable a machine to understand a video story is to train the machine to answer questions about the video story~\cite{schank2013scripts,mueller2004understanding}.
Recently, several video question answering (Video QA) datasets~\cite{tapaswi2015,kim2017,mun2017marioQA,jang2017,lei2018tvqa} have been released publicly. 
These datasets encourage inspiring works in this domain, but they do not give sufficiently careful consideration of some important aspects of video story understanding.  
Video QA datasets can be used not only for developing video story understanding models but also for evaluating the degree of intelligence of the models. Therefore, QAs should be collected considering difficulty levels of the questions to evaluate the degree of story understanding intelligence~\cite{collis1975study}. 
However, the collected QAs in the previous studies are highly-biased and lack of variance in the levels of question difficulty. 
Furthermore, while focalizing on characters within a story is important to form local story coherence~\cite{riedl2010narrative,grosz1995centering}, previous works did not provide any consistent annotations for characters to model this coherence.

In this work, we propose a new Video QA task, DramaQA, for a more comprehensive understanding of the video story.
1) We focus on the \textit{understanding} with hierarchical QAs used as a hierarchical evaluation metric based on the cognitive-developmental stages of human intelligence.
We define the level of understanding in conjunction with Piaget's theory~\cite{collis1975study} and collect QAs accordingly. In accordance with~\cite{yjheo2019}, we collect questions along with one of four hierarchical difficulty levels, based on two criteria; memory capacity (MC) and logical complexity (LC).
With these hierarchical QAs, we offer a more sophisticated evaluation metric to measure understanding levels of Video QA models.
2) We focus on the \textit{story} with character-centered video annotations.
To learn character-centered video representations, the DramaQA provides rich annotations for main characters such as visual bounding boxes, behaviors and emotions of main characters and also coreference resolved scripts. 
By sharing character names for all the annotations including QAs, the model can have a coherent view of characters in the video story.
3) We provide {\modelname} model to answer the questions for the multimodal story by utilizing the character-centered annotations. Using representations of two different abstraction levels, our model hierarchically learns underlying correlations between the video clips, QAs, and characters.


\section{Related Work}
This section introduces Question and Answering about Story and Cognitive Developmental Stages of Humans. Because of the page limit, we inroduce Video Understanding in appendix A.
 \subsection{Question and Answering about Story}
 Question and answering (QA) has been commonly used to evaluate reading comprehension ability of textual story. 
 \cite{hermann2015teaching,trischler2016newsqa} introduced QAs dataset about news articles or daily emails, and \cite{richardson2013mctest,hill2015goldilocks} dealt with QAs built on children's book stories. Especially, NarrativeQA suggested by \cite{kovcisky2018narrativeqa} aims to understand the underlying narrative about the events and relations across the whole story in book and movie scripts, not the fragmentary event. \cite{mostafazadeh2016corpus} established ROCStories capturing a set of causal and temporal relations between daily events, and suggested a new commonsense reasoning framework `Story Cloze Test' for evaluating story understanding.
 
 Over the past years, increasing attention has focused on understanding of multimodal story, not a textual story. By exploiting multimodalities, the story delivers the more richer semantics without ambiguity. \cite{tapaswi2015,kim2017,mun2017marioQA,jang2017,lei2018tvqa} considered the video story QA task as an effective tool for multimodal story understanding and built video QA datsets. The more details on the comparison of those datasets with the proposed dataset, DramaQA, are described in the section titled `Comparison with Other Video QA Datasets.'


\begin{figure*}[!h]
 \centering
 \includegraphics[width=1.0\textwidth]{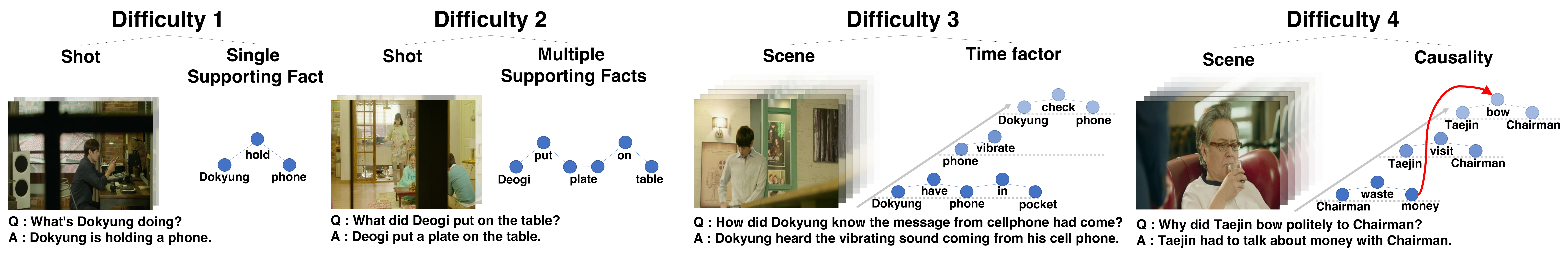}
 \caption{Four examples of different QA level. Difficulty 1 and 2 target shot-length videos. Difficulty 1 requires single supporting fact to answer, and Difficulty 2 requires multiple supporting facts to answer. Difficulty 3 and 4 require a time factor to answer and target scene-length videos. Especially, Difficulty 4 requires causality between supporting facts from different time.}
 \label{fig:multilevelqa}
 \end{figure*}
 \index{figure}
 
\subsection{Cognitive Developmental Stages of Humans}
 
 We briefly review the cognitive development of humans based on Piaget’s theory~\citep{piaget1972intellectual,collis1975study} that is a theoretical basis of our proposed hierarchical evaluation metric for video story understanding. Piaget's theory explains in detail the developmental process of human cognitive ability in conjunction with information processing. 

 At \textit{Pre-Operational Stage} (4 to 6 years), a child thinks at a symbolic level, but is not yet using cognitive operations. The child can not transform, combine or separate ideas. Thinking at this stage is not logical and often unreasonable. At  \textit{Early Concrete Stage} (7 to 9 years), a child can utilize only one relevant operation. Thinking at this stage has become detached from instant impressions and is structured around a single mental operation, which is a first step towards logical thinking. At  \textit{Middle Concrete Stage} (10 to 12 years), a child can think by utilizing more than two relevant cognitive operations and acquire the facts of dialogues. This is regarded as the foundation of proper logical functioning. However, the child at this stage lacks own ability to identify general fact that integrates relevant facts into coherent one. At  \textit{Concrete Genelization Stage} (13 to 15 years), a child can think abstractly, but just generalize only from personal and concrete experiences. The child does not have own ability to hypothesize possible concepts or knowledge that is quite abstract.  \textit{Formal Stage} (16 years onward) is characterized purely by abstract thought. Rules can be integrated to obtain novel results that are beyond the individual’s own personal experiences. In this paper, we carefully design the evaluation metric for video story understanding with the question-answer hierarchy for levels of difficulty based on the cognitive developmental stages of humans.

 \begin{figure*}[t]
 \centering
 \includegraphics[width=1.0\textwidth]{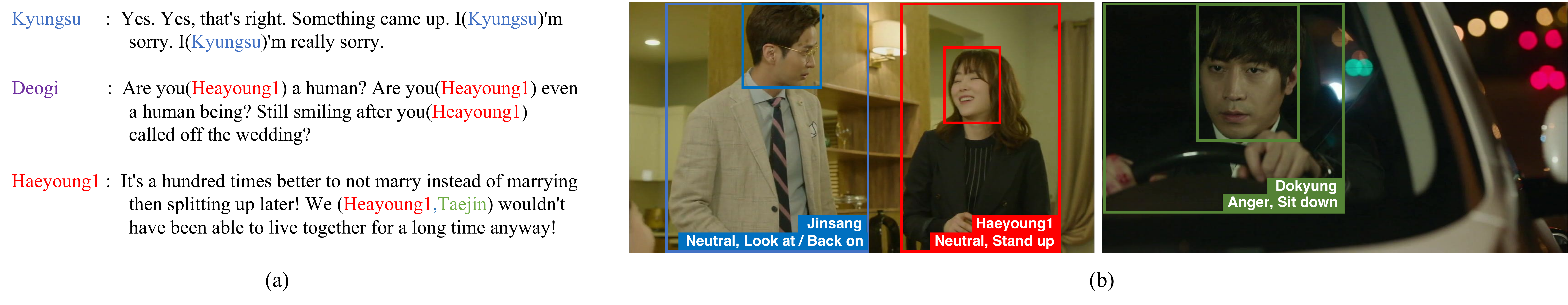}
 \caption{Examples of character-centered video annotations: (a) coreference resolved scripts and (b) visual metadata which contains the main characters' bounding box, name, behavior, and emotion. All annotations for characters in script and visual metadata can be co-referred by unique character's name.} 
 
 \label{fig:visual}
 \end{figure*}
 \vspace{-0.5em}
 \index{figure} 

\section{DramaQA Dataset}
 
We collect the dataset on a popular Korean drama \textit{Another Miss Oh}, which has 18 episodes, {\vidhr} hours in total. DramaQA dataset contains {\clips} various length video clips which consist of sequences of video frames (3 frames per second) and {\qa} multiple choice QA pairs with hierarchical difficulty levels.
Furthermore, it includes rich character-centered annotations such as visual bounding boxes, behaviors and emotions of main characters, and coreference resolved scripts. 
Figure~\ref{fig:dramaqa_overview} illustrates the DramaQA dataset. 
Also, detailed information of the dataset including various attributes, statistics and collecting procedure can be found in Appendix B.
 

\subsection{Question-Answer Hierarchy for Levels of Difficulty}\label{sec:3.2}
 
To collect question-answer pairs with levels of difficulty, 
we propose two criteria: Memory capacity and logical complexity.
Memory capacity (MC) is defined as the required length of the video clip to answer the question, and corresponds to working memory in human cognitive process.
Logical complexity (LC) is defined by the number of logical reasoning steps required to answer the question, which is in line with the hierarchical stages of human development~\cite{seol2011stanford}. 

\begin{table*}[ht]
\begin{threeparttable}
\begin{tabular}{l c c c c c c} 
\specialrule{.1em}{.05em}{.05em} 
     & \# Q  
     & \begin{tabular}{@{}c@{}}\# Annotated \\ Images\end{tabular}
     & \begin{tabular}{@{}c@{}}Avg. Video \\ len. (s)\end{tabular}
     & \begin{tabular}{@{}c@{}}Textual \\ metadata\end{tabular}
     & \begin{tabular}{@{}c@{}}Visual \\ metadata\end{tabular}
     & Q. lev \\
\specialrule{.1em}{.05em}{.05em} 
    TGIF-QA~\cite{jang2017}           & 165,165   & -    & 3.1   & - & - & -\\ 
    MarioQA~\cite{mun2017marioQA}    & 187,757   & -       & $<$ 6     & - & - & -\\
    PororoQA~\cite{kim2017}              & 8,913     & -     & 1.4   & \begin{tabular}{@{}c@{}} Description,  \\ Subtitle \end{tabular}& -& -\\
    MovieQA~\cite{tapaswi2015}               & 6,462     & -     & 202.7 & \begin{tabular}{@{}c@{}} Plot, DVS, \\ Subtitle \end{tabular}  & -  & -\\
    TVQA~\cite{lei2018tvqa}             & 152,545   & -   & 76.2  & Script  & - & -\\ 
    TVQA+~\cite{lei2019tvqa+}& 29,383 & 148,468 & 61.49 & Script & Char./Obj. Bbox\tnote{**} & \\
    \hline
    DramaQA       & {\qa}     & {\visual} & \begin{tabular}{@{}c@{}}3.7\tnote{a}\\91.3\tnote{b}\end{tabular}     & Script\tnote{*} & \begin{tabular}{@{}c@{}} Char. Bbox,  \\ Behavior, Emotion \end{tabular} &\checkmark \\ 
\specialrule{.1em}{.05em}{.05em} 
\end{tabular}
\begin{tablenotes}[para]
\item[a] Avg. video length for shot
\item[b] Avg. video length for scene
\item[*] Coreference resolved script
\item[**] Only mentioned in QAs
\end{tablenotes}
\end{threeparttable}
\centering
\caption{Comparison between video story QA datasets. Only DramaQA dataset provides hierarchical QAs from shot-level and scene-level videos and character-centered visual metadata (bounding box, name, behavior, and emotion).}
\label{table:dataset}
\vspace{-1.0em}
\end{table*}

\subsubsection{Criterion 1: Memory Capacity}
The capacity of working memory increases gradually over childhood, as does cognitive and reasoning ability required for higher level responses~\cite{case1980implications,mclaughlin1963psycho,pascual1969cognitive}. 
In the Video QA problem, the longer video story to answer a question requires, the harder to reason the answer from the video story is. 
Here, we consider two levels of memory capacity; shot and scene. 
Detailed definitions of each level are below:

\begin{itemize}[label=$\bullet$]
\item \textbf{Level 1 (shot):} The questions for this level are based on video clips mostly less than about 10 seconds long, shot from a single camera angle. This set of questions can contain atomic or functional/meaningful action in the video. Many Video QA 
datasets 
belong to this level~\cite{jang2017,maharaj2017,mun2017marioQA, kim2017}. 

\item \textbf{Level 2 (scene):} The questions for this level are based on clips that are about 1-10 minutes long without changing location. Videos at this level contain sequences of actions, which augment the shots from Level 1. We consider this level as the ``story'' level according to our working definition of story. MovieQA~\cite{tapaswi2015} and TVQA+~\cite{lei2019tvqa+} are the only datasets which belong to this level. 
\end{itemize}

\subsubsection{Criterion 2: Logical Complexity}
Complicated questions often require more (or higher) logical reasoning steps than simple questions. 
In a similar vein, if a question needs only a single supporting fact with single relevant datum, we regard this question as having low logical complexity. 
Here, we define four levels of logical complexity from simple recall to high-level reasoning, similar to hierarchical stages of human development~\cite{seol2011stanford}.
\begin{itemize}[label=$\bullet$]
    \item \textbf{Level 1 (Simple recall on one cue):} The questions at this level can be answered using simple recall; requiring only one supporting fact. Supporting facts are represented as triplets in form of \textit{\{subject-relationship-object\}} such as \textit{\{person-hold-cup\}}. 
    
    \item \textbf{Level 2 (Simple analysis on multiple cues):} 
    These questions require recall of multiple supporting facts, which trigger simple inference. For example, two supporting facts \textit{\{tom-in-kitchen\}} and \textit{\{tom-grab-tissue\}} are referenced to answer ``Where does Tom grab the tissue?". 
    
    \item \textbf{Level 3 (Intermediate cognition on dependent multiple cues):} The questions at this level require multiple supporting facts with time factor to answer. 
    Accordingly, the questions at this level cover how situations have changed and subjects have acted. 

    \item \textbf{Level 4 (High-level reasoning for causality):} The questions at this level cover reasoning for causality which can begin with ``Why''. Reasoning for causality is the process of identifying causality, which is the relationship between cause and effect from actions or situations. 
\end{itemize}

\subsubsection{Hierarchical Difficulties of QA aligned with Cognitive Developmental Stages}

From the two criteria, we define four hierarchical difficulties for QA which are consistent with cognitive developmental stages of Piaget's theory~\cite{piaget1972intellectual,collis1975study}. 
Questions at level 1 in MC and LC belong to Difficulty 1 which is available from \textit{Pre-Operational Stage} where a child thinks at a symbolic level, but is not yet using cognitive operations. Questions at level 1 in MC and level 2 in LC belong to Difficulty 2 which is also available from \textit{Early Concrete Stage} where a child can utilize a relevant operation between multiple supporting facts. 
Questions at level 2 in MC and level 3 in LC belong to Difficulty 3 which is available from \textit{Middle Concrete Stage} where a child can think by utilizing more than two relevant cognitive operations and utilize dependent multiple supporting facts across time. 
Questions at level 2 in MC and level 4 in LC belong to Difficulty 4 which is available from \textit{Concrete Generalization Stage} where a child can just generalize only from personal and concrete experience and have a higher thought on causality in relation to ``Why".
Examples for each Difficulty are illustrated in Figure~\ref{fig:multilevelqa}.



\subsection{Character-Centered Video Annotations}
\label{sec:3.3}
As the characters are primary components of stories, we provide rich annotations for the main characters in the video contents. 
As visual metadata, main characters are localized in the appeared image frames sampled in video clips and annotated with not only the character names but also behavior and emotion states.
Also, all coreferences (e.g. he/she/they) of the main characters in scripts are resolved to give a consistent view of the characters.
Figure~\ref{fig:visual} shows the examples of visual metadata and coreference resolved scripts.

\subsubsection{Visual Metadata}

 \begin{itemize}[label=$\bullet$]
 \item \textbf{Bounding Box: }In each image frame, bounding boxes of both a face rectangle and a full-body rectangle for the main characters are annotated with their name. In total, 20 main characters are annotated with their unique name. 
 \item \textbf{Behavior \& Emotion: }Along with bounding boxes, behaviors and emotions of the characters shown in the image frames are annotated.
 Including \texttt{none} behavior, total 28 behavioral verbs, such as \texttt{drink, hold, cook}, are used for behavior expression. 
 Also, we present characters' emotion with 7 emotional adjectives; \texttt{anger}, \texttt{disgust}, \texttt{fear}, \texttt{happiness}, \texttt{sadness}, \texttt{surprise}, and \texttt{neutral}.
 
 \end{itemize}
 
 


 

\subsubsection{Coreference Resolved Scripts}
To understand video stories, especially drama, it is crucial to understand the dialogue between the characters.
Notably, the information such as ``\textit{Who} is talking to \textit{whom} about \textit{who} did what?'' is significant in order to understand whole stories.
In DramaQA, we provide this information by resolving all coreferences for main characters in scripts.
As shown in Figure~\ref{fig:visual}(a), we annotate the characters' names to all personal pronouns for characters, such as I, you, we, him, etc.
By doing so, characters in scripts can be matched with those in visual metadata and QAs.

\begin{figure*}[!h]
 \centering
 \includegraphics[width=1.0\textwidth]{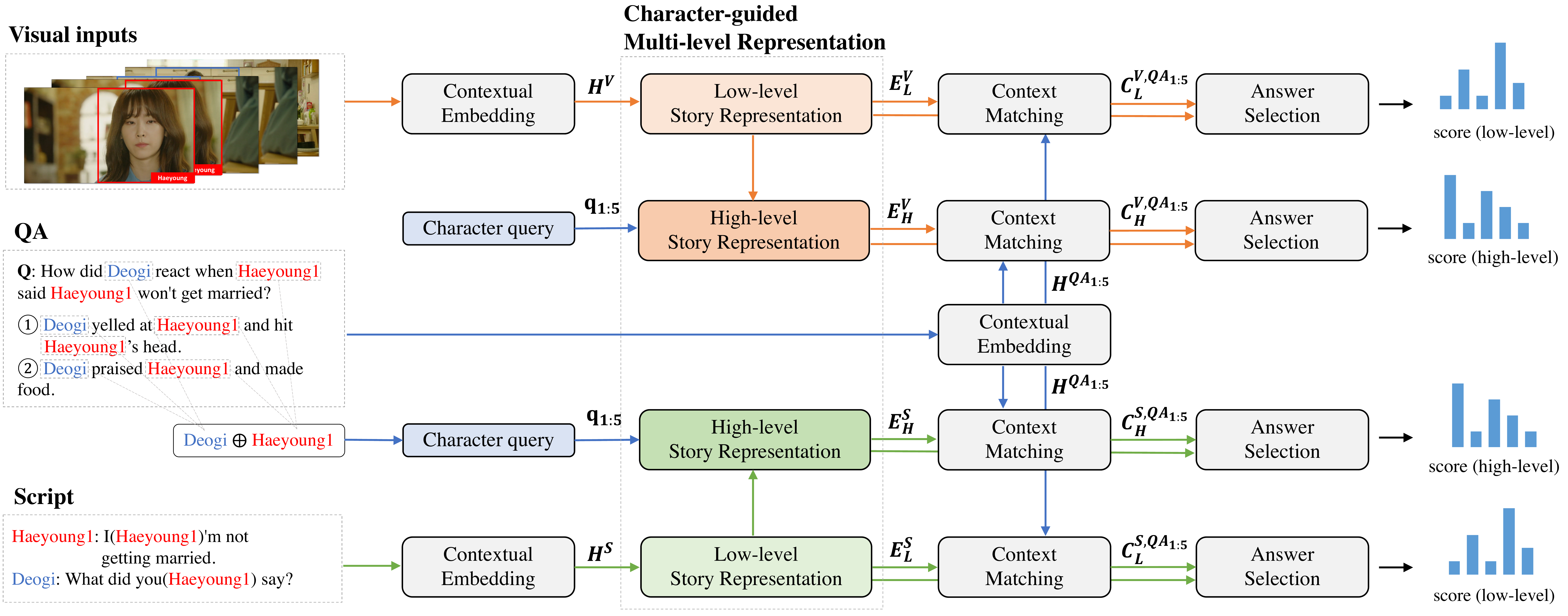}
 \caption{Our {\modelname} model learns underlying correlations between the video clips, QAs, and characters using low-level and high-level representations. Final score for answer selection is the sum of each input stream's output score.}
 \label{fig:model}
 \end{figure*}
\vspace{-1.0em}
 
 \index{figure}

\subsection{Comparison with Other Video QA Datasets}\label{comp}

We also present a comparison of our dataset to some recently proposed video QA datasets (Table~\ref{table:dataset}). TGIF-QA and MarioQA~\cite{jang2017,mun2017marioQA} only dealt with a sequence of images not textual metadata, and focused on spatio-temporal reasoning tasks about short video clips. PororoQA~\cite{kim2017} was created using animation videos that include simple stories that happened in a small closed world. Since most of the questions in PororoQA are very relevent to subtitles and descriptions, most answers can be solved only using the textual information. MovieQA~\cite{tapaswi2015} contains movie clips and various textual metadata such as plots, DVS, and subtitles. However, since the QA pairs were created based on plot synopsis without watching the video, collected questions are not grounded well to the video contents. TVQA+~\cite{lei2019tvqa+}, a sequel to the TVQA~\cite{lei2018tvqa} particularly included annotated images with bounding boxes linked with characters and objects only mentioned in QAs. Although TVQA+ provides spatial and temporal annotations for answering a given question, most of their questions are aligned to relatively short moments (less than 15 seconds). 
Among the datasets, only the DramaQA 1) provides difficulty levels of the questions and rich information of characters including visual metadata and coreference resolved scripts and 2) tackles both shot-level and scene-level video clips.

\begin{table*}[h!]

\begin{tabular}{l @{\hskip 0.2in} c @{\hskip 0.2in} c @{\hskip 0.2in}c @{\hskip 0.2in} c @{\hskip 0.1in} | @{\hskip 0.1in} c @{\hskip 0.1in} c @{\hskip 0.1in} c } 
\specialrule{.1em}{.05em}{.05em} 
    Model
     & Diff. 1
     & Diff. 2
     & Diff. 3 
     & Diff. 4
     & Overall 
     & Diff. Avg.\\
\specialrule{.1em}{.05em}{.05em} 
    QA Similarity & 30.64   & 27.20   & 26.16   & 22.25    & 28.27   & 26.56  \\
    \hline
    S.Only$-$Coref     & 54.43  & 51.19  & 49.71  & 52.89  & 52.89  & 52.06  \\
    S.Only     & 62.03   & 63.58  & 56.15  & 55.58  & 60.95  & 59.34  \\
    V.Only$-$V.Meta     & 63.28   & 56.86  & 49.88  & 54.44  & 59.06  & 56.11  \\
    V.Only     & 74.82   & 70.61  & 54.60  & 56.48  & 69.22  & 64.13  \\
    \hline
    Our$-$High  & 75.68  & 72.53  & 54.52  & 55.66  & 70.03  & 64.60  \\
    Our$-$Low   & 74.49  & 72.37  & 55.26  & 56.89   & 69.60  & 64.75  \\
    \hline
    Our (Full)  & 75.96  & 74.65  & 57.36  & 56.63  & 71.14   & 66.15   \\
    
\specialrule{.1em}{.05em}{.05em} 
\end{tabular}
\centering
\caption{Quantitative result for our model on test split. Last two columns show the performance of overall test split and the average performance of each set. \texttt{S.Only} and \texttt{V.Only} indicate our model only with script and visual inputs respectively. \texttt{S.Only$-$Coref.} and \texttt{V.Only$-$V.Meta} are \texttt{S.Only} with removed coreference and speaker annotation and \texttt{V.Only} with removed visual metadata. \texttt{Our(Full)} contains all elements of our model. \texttt{Our$-$High} and \texttt{Our$-$Low} are with removed high-level representations and with remove low-level representations from \texttt{Our(Full)}.}
\label{table:quan_result}
\vspace{-1.0em}
\end{table*}

\section{Model}
We propose {\modelname} model which grounds evidence in coherent characters to answer questions about the video. Our main goal is to build a QA model that hierarchically understands the multimodal story, by utilizing the character-centered annotations. The proposed model consists of two streams (for vision and textual modality) and multi-level (low and high) for each stream. The low-level representations imply the context of the input stream with annotations related to main characters. From low-level representations, we get high-level representations using character query appeared in QA. Then we use Context Matching module to get a QA-aware sequence for each level. Outputs of these squences are converted to a score for each answer candidate to select the most appropriate answer. Figure \ref{fig:model} shows our network architecture.\footnote{https://github.com/liveseongho/DramaQA}

 \subsection{Contextual Embedding Module}
 \label{sec:4.1}
 An input into our model consists of a question, a set of five candidate answers, and two types of streams related to video context which are coreference resolved scripts and visual metadata.
 Each question is concatenated to its five corresponding answer candidates. We denote a QA pair as $\text{QA}_i \in \mathbb{R}^{(T_Q+T_{A_i}) \times D_W}$, where $T_Q$ and $T_{A_i}$ are the length of each sentence and $D_W$ is the word embedding dimension. 
 We denote the input stream from the script $S \in \mathbb{R}^{T_{\text{sent}} \times T_{\text{word}} \times D_W}$ where $T_{\text{sent}}$ is the number of setences and $T_{\text{word}}$ is the maximum number of words per a sentence.
 Behavior and emotion are converted to word embedding and concatenated to each bounding box feature. We denote the visual metadata stream $V \in \mathbb{R}^{T_{\text{shot}} \times T_{\text{frame}} \times (D_V+2*D_W)}$ where $T_{\text{shot}}$ is the number of shots in clips, $T_{\text{frame}}$ is the number of frames per a shot, and $D_V$ is the feature dimension of each bounding box.
 
 In order to capture the coherence of characters, we also use a speaker of script and a character's name annotated in bounding box. Both pieces of character information are converted to one-hot vector and concatenated to input streams respectively.
 Then, we use bi-directional LSTM to get streams with temporal context from input streams, and we get $H^{\text{QA}_i}\in \mathbb{R}^{(T_Q+T_{A_i}) \times D}$, $H^S\in \mathbb{R}^{T_{\text{sent}} \times T_{\text{word}} \times D}$ and $H^V\in \mathbb{R}^{T_{\text{shot}} \times T_{\text{frame}} \times D}$ for each stream, respectively. 
 
 
 \subsection{Character-guided Multi-level Representation}
 \label{sec:4.2}
 Under the assumption that there is background knowledge that covers the entire video clip, such as the characteristics of each of the main characters, we have global representations for each character name $\mathbf{m} \in \mathbb{R}^{d}$, where $d$ is a dimension of each character representation. In our case $d$ is same with the dimension of each contextual embedding. We use characters in question and $i$-th candidate answer pair to get character query $\mathbf{q}_i = \sum_j \mathbf{m}_j$. 
 
 Using this $\mathbf{q}_i$ as a query, we obtain character-guided high-level story representations for each stream $E_H^V$ and $E_H^S$ from low-level contextual embeddings by using attention mechanism:
 \begin{equation}
    E_H^V[j]=\text{softmax}(\mathbf{q}_i H^V[j]^\top)H^V[j]
 \end{equation}
 \begin{equation}
    E_H^S[j]=\text{softmax}(\mathbf{q}_i H^S[j]^\top)H^S[j]
 \end{equation}
 We note that $E_H^V[j]$ and $E_H^S[j]$ represent sentence-level embedding for script and shot-level embedding for visual inputs, respectively.
 For the low-level story representations, we flatten $H^S$ and $H^V$ to 2-D matrices, so that $E_L^S\in\mathbb{R}^{(T_{\text{sent}}*T_{\text{word}})\times D}$ and $E_L^V\in\mathbb{R}^{(T_{\text{shot}}*T_{\text{frame}})\times D}$ is obtained.
 
 \subsection{Context Matching Module}
 \label{sec:4.3}
 The context matching module converts each input sequence to a query-aware context by using the question and answers as a query. 
 This approach was taken from attention flow layer in \cite{Seo2016BidirectionalAF, lei2018tvqa}. 
 Context vectors are updated with a weighted sum of query sequences based on the similarity score between each query timestep and its corresponding context vector. We can get  $C^{S,{\text{QA}_i}}$ from $E^{S}$ and $C^{V,{\text{QA}_i}}$ from $E^{V}$.

 \subsection{Answer Selection Module}
 \label{sec:4.4}
 For embeddings of each level from script and visual inputs, we concatenate $E^S$, $C^{S,\text{QA}_i}$, and $E^S \odot C^{S,\text{QA}_i}$, where $\odot$ is the element-wise multiplication. 
 We also concatenate boolean flag $f$ which is \texttt{TRUE} when the speaker or the character name in script and visual metadata appears in the question and answer pair. 
 
 \begin{equation}
    X_L^{S_i} = [E_L^S ; C_L^{S,\text{QA}_i} ; E_L^S \odot C_L^{S,\text{QA}_i} ; f]
 \end{equation}
 \begin{equation}
    X_H^{S_i} = [E_H^S ; C_H^{S,\text{QA}_i} ; E_H^S \odot C_H^{S,\text{QA}_i} ; f]
 \end{equation}
  \noindent where we can get $ X_L^{V_i}$ and $ X_H^{V_i}$ in the same manner.
 
 For each stream $ X_L^{S_i}$, $ X_H^{S_i}$, $ X_L^{V_i}$, $ X_H^{V_i}$, we apply 1-D convolution filters with various kernel sizes and concatenate the outputs of the kernels. 
 Applying max-pool over time and linear layer, we calculate scalar score for $i$-th candidate answer. The final output score is simply the sum of output scores from the four different streams, and the model selects the answer candidate with the largest final output score as the correct answer.

\begin{figure*}[!h]
 \centering
 \includegraphics[width=0.95\textwidth]{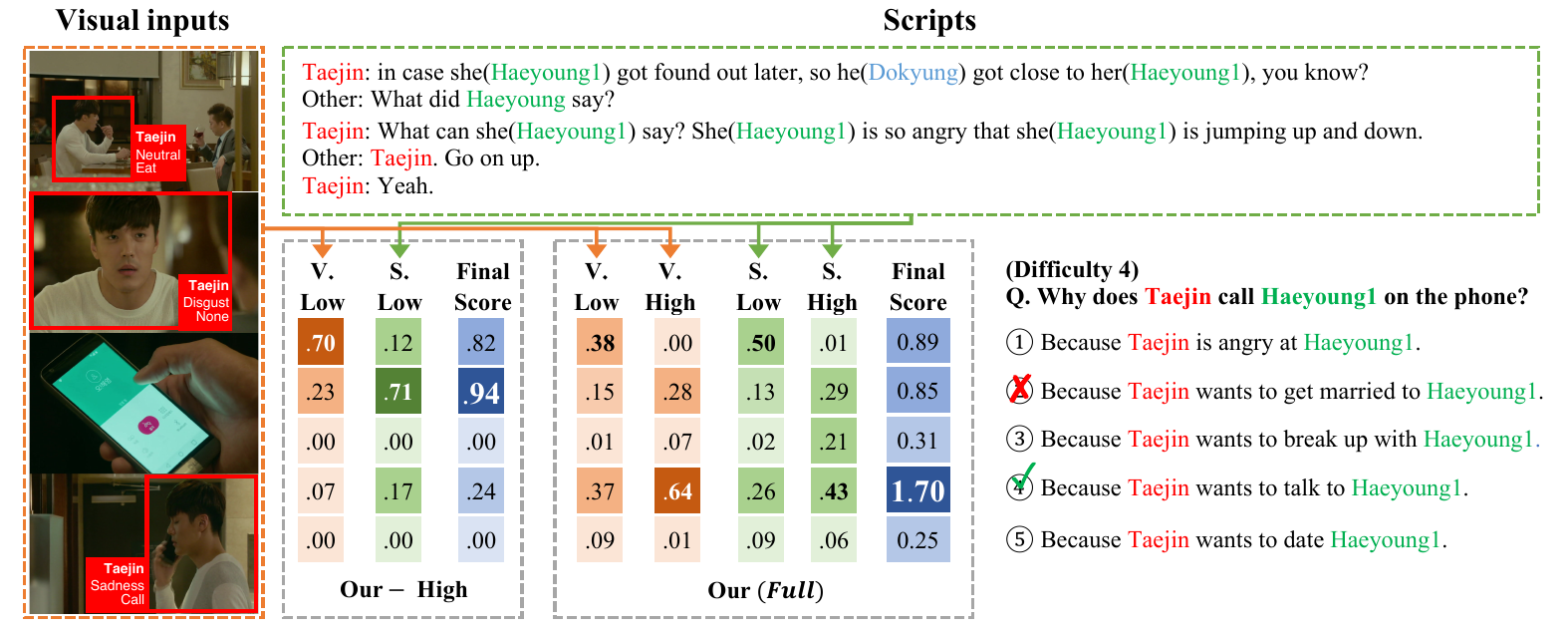}
 \caption{
 An example of correct prediction case to answer the question in Difficulty 4. To see the effectiveness of multi-level representation, we present the results of \texttt{Our(Full)} and \texttt{Our$-$High} in parallel. Scores of visual inputs are colored in orange and scores of scripts are colored in green. We colored final scores in blue. Prediction of \texttt{Our(Full)} is indicated by green checkmark which is ground truth answer, and prediction of \texttt{Our$-$High} is indicated by red crossmark.}
 \label{fig:qual_result}
 \vspace{-1.0em}
 \end{figure*}
 \index{figure}

\section{Results}
 \subsection{Quantitative Results}

Here, we discuss an ablation study to analyze the model's characterstics profoundly. 
 Table \ref{table:quan_result} shows the quantitative results of the ablation study for our model, and we described our experimental settings and implementation details in the Appendix C. \texttt{QA Similarity} is a simple baseline model designed to choose the highest score on the cosine similarity between the average of question's word embeddings and the average of candidate answer's word embeddings. The overall test accuracy of \texttt{Our(Full)} was 71.14\% but the performance of each difficulty level varies. The tendency of poor performance as the level of difficulty increases shows that the proposed evaluation criteria considering the cognitive developmental stages are designed properly.


 
 To confirm the utilization of multi-level architecture is effective, we compare the performance of our full model \texttt{Our(Full)} with those of the model excluding the high-level story representation module \texttt{Our$-$High} and the model excluding the low-level story representation module \texttt{Our$-$Low}.
 We can see that performances on Diff. 3 and 4 are more degraded in \texttt{Our$-$High} than \texttt{Our$-$Low}, whereas performances on Diff. 1 and 2 are more degraded \texttt{Our$-$Low} than \texttt{Our$-$High}.
 These experimental results indicate that the high-level representation module helps to handle difficult questions whereas the low-level representation module is useful to model easy questions.

 
  
 Note that both script and visual input streams are helpful to infer a correct answer. \texttt{S.Only} uses only the script as the input and shows a sharp decline for Diff. 1 and 2. Since about 50\% of QAs at Diff. 1 and 2 has a (shot-level) target video without a script, such questions need to be answered only with visual information.
 \texttt{V.Only} uses only visual input and shows decent performance on the overall difficulties. 
 Especially, the results show that the rich visual information is dominantly useful to answer the question at Diff. 1 and 2.
 
 To check the effectiveness of character-centered annotation, we experimented with two cases: \texttt{V.Only$-$V.Meta} and \texttt{S.Only$-$Coref}. Here, \texttt{V.Only$-$V.Meta} only includes the visual feature of the corresponding frame by excluding visual metadata (bounding box, behavior, and emotion) of the main characters. Since it is hard to exactly match between characters of QA and video frames, the performance of \texttt{V.Only$-$V.Meta} was strictly decreased. For the same reason, \texttt{S.Only-Coref}, which removed coreferences and speakers from the \texttt{S.Only}, showed low performance in overall.
 These results show the effect of the proposed approach on character-centered story understanding.
 
 We also compared our model with recently proposed methods for other video QA datasets. Due to the space limitation, the results are described in the Appendix D. 
 
\subsection{Qualitative Results}
In this section, we demonstrate how each module of the proposed model works to answer questions.
As shown in Figure \ref{fig:qual_result}, our model successfully predicts an answer by matching the context from candidate answers with the context from each input source. Especially, it shows that high-level representations help to infer a more appropriate answer from the context. In \texttt{Our(Full)}, Low-level scores from scripts of our model confused the answer with the first candidate including the word \textit{angry}, but high-level scores from scripts chose the ground truth answer. Also, low-level scores from visual inputs inferred the first candidate answer to be correct based on the visual metadata \textit{disgust, sadness}, but high-level scores from visual inputs gave more weight to the fourth candidate answer. 
As we discussed, character-guided high-level representations help to answer the question which requires complex reasoning.
Without the high-level representations (shown in the results of \texttt{Our$-$High}), the model cannot fully understand the story and focuses on the low-level details.
More examples including failures are provided in the Appendix E.

\section{Conclusion}
To develop video story understanding intelligence, we propose DramaQA dataset. Our dataset has cognitive-based difficulty levels for QA as a hierarchical evaluation metric. Also, it provides coreference resolved script and rich visual metadata for character-centered video. We suggest a {\modelname} model to verify the usefulness of multi-level modeling and character-centered annotation. Using both low-level and high-level representations, our model efficiently learns underlying correlations between the video clips, QAs and characters.

The application area of the proposed DramaQA dataset is not limited to QA based video story understanding. Our DramaQA dataset with enriched metadata can be utilized as a good resource for video-related researches including emotion or behavior analysis of characters, automatic coreference identification from scripts, and coreference resolution for visual-linguistic domain. Also, our model can be utilized as a fine starting point for resolving the intrinsic challenges in the video story understanding such as the integrated multimodal data analysis.


As future work, we will extend the two criteria of hierarchical QA so that the dataset can deal with longer and more complex video story along with expanding the coverage of evaluation metric. Also, we plan to provide hierarchical character-centered story descriptions, objects, and places. We expect that our work can encourage inspiring works in the video story understanding domain.

\section*{Acknowledgement}
This work was partly supported by the Institute for Information \& Communications Technology Promotion (2015-0-00310-SW.StarLab, 2017-0-01772-VTT, 2018-0-00622-RMI, 2019-0-01367-BabyMind) and Korea Institute for Advancement Technology (P0006720-GENKO) grant funded by the Korea government.

\bigskip

\bibliography{aaai21}

\begin{thebibliography}{50}
\providecommand{\natexlab}[1]{#1}
\providecommand{\url}[1]{\texttt{#1}}
\providecommand{\urlprefix}{URL }
\expandafter\ifx\csname urlstyle\endcsname\relax
  \providecommand{\doi}[1]{doi:\discretionary{}{}{}#1}\else
  \providecommand{\doi}{doi:\discretionary{}{}{}\begingroup
  \urlstyle{rm}\Url}\fi

\bibitem[{Abu-El-Haija et~al.(2016)Abu-El-Haija, Kothari, Lee, Natsev,
  Toderici, Varadarajan, and Vijayanarasimhan}]{abu2016youtube}
Abu-El-Haija, S.; Kothari, N.; Lee, J.; Natsev, P.; Toderici, G.; Varadarajan,
  B.; and Vijayanarasimhan, S. 2016.
\newblock Youtube-8m: A large-scale video classification benchmark.
\newblock \emph{arXiv preprint arXiv:1609.08675} .

\bibitem[{Caba~Heilbron et~al.(2015)Caba~Heilbron, Escorcia, Ghanem, and
  Carlos~Niebles}]{caba2015activitynet}
Caba~Heilbron, F.; Escorcia, V.; Ghanem, B.; and Carlos~Niebles, J. 2015.
\newblock Activitynet: A large-scale video benchmark for human activity
  understanding.
\newblock In \emph{Proceedings of the ieee conference on computer vision and
  pattern recognition}, 961--970.

\bibitem[{Case(1980)}]{case1980implications}
Case, R. 1980.
\newblock Implications of neo-Piagetian theory for improving the design of
  instruction.
\newblock \emph{Cognition, development, and instruction} 161--186.

\bibitem[{Collis(1975)}]{collis1975study}
Collis, K.~F. 1975.
\newblock \emph{A Study of Concrete and Formal Operations in School
  Mathematics: A Piagetian Viewpoint.}
\newblock Hawthorn Vic : Australian Council for Educational Research.

\bibitem[{Devlin et~al.(2018)Devlin, Chang, Lee, and
  Toutanova}]{devlin2018pretraining}
Devlin, J.; Chang, M.-W.; Lee, K.; and Toutanova, K. 2018.
\newblock BERT: Pre-training of Deep Bidirectional Transformers for Language
  Understanding.
\newblock \urlprefix\url{http://arxiv.org/abs/1810.04805}.
\newblock Cite arxiv:1810.04805Comment: 13 pages.

\bibitem[{Gao et~al.(2019)Gao, Zeng, Song, Li, Liu, Mei, and Shen}]{gao2019}
Gao, L.; Zeng, P.; Song, J.; Li, Y.-F.; Liu, W.; Mei, T.; and Shen, H.~T. 2019.
\newblock Structured Two-Stream Attention Network for Video Question Answering.
\newblock \emph{Proceedings of the AAAI Conference on Artificial Intelligence}
  33(01): 6391--6398.
\newblock \doi{10.1609/aaai.v33i01.33016391}.
\newblock
  \urlprefix\url{https://ojs.aaai.org/index.php/AAAI/article/view/4602}.

\bibitem[{Girdhar and Ramanan(2020)}]{girdhar2020cater}
Girdhar, R.; and Ramanan, D. 2020.
\newblock {CATER: A diagnostic dataset for Compositional Actions and TEmporal
  Reasoning}.
\newblock In \emph{ICLR}.

\bibitem[{Grosz, Weinstein, and Joshi(1995)}]{grosz1995centering}
Grosz, B.~J.; Weinstein, S.; and Joshi, A.~K. 1995.
\newblock Centering: A framework for modeling the local coherence of discourse.
\newblock \emph{Computational linguistics} 21(2): 203--225.

\bibitem[{He et~al.(2015)He, Zhang, Ren, and Sun}]{He2015DeepRL}
He, K.; Zhang, X.; Ren, S.; and Sun, J. 2015.
\newblock Deep Residual Learning for Image Recognition.
\newblock \emph{2016 IEEE Conference on Computer Vision and Pattern Recognition
  (CVPR)} 770--778.

\bibitem[{Heo et~al.(2019)Heo, On, Choi, Lim, Kim, Ryu, Bae, and
  Zhang}]{yjheo2019}
Heo, Y.; On, K.; Choi, S.; Lim, J.; Kim, J.; Ryu, J.; Bae, B.; and Zhang, B.
  2019.
\newblock Constructing Hierarchical Q{\&}A Datasets for Video Story
  Understanding.
\newblock \emph{CoRR} abs/1904.00623.
\newblock \urlprefix\url{http://arxiv.org/abs/1904.00623}.

\bibitem[{Hermann et~al.(2015)Hermann, Kocisky, Grefenstette, Espeholt, Kay,
  Suleyman, and Blunsom}]{hermann2015teaching}
Hermann, K.~M.; Kocisky, T.; Grefenstette, E.; Espeholt, L.; Kay, W.; Suleyman,
  M.; and Blunsom, P. 2015.
\newblock Teaching machines to read and comprehend.
\newblock In \emph{Advances in neural information processing systems},
  1693--1701.

\bibitem[{Hill et~al.(2016)Hill, Bordes, Chopra, and
  Weston}]{hill2015goldilocks}
Hill, F.; Bordes, A.; Chopra, S.; and Weston, J. 2016.
\newblock The Goldilocks Principle: Reading Children's Books with Explicit
  Memory Representations.
\newblock In Bengio, Y.; and LeCun, Y., eds., \emph{4th International
  Conference on Learning Representations, {ICLR} 2016, San Juan, Puerto Rico,
  May 2-4, 2016, Conference Track Proceedings}.
\newblock \urlprefix\url{http://arxiv.org/abs/1511.02301}.

\bibitem[{Huang et~al.(2020)Huang, Chen, Zeng, Du, Tan, and Gan}]{huang2020}
Huang, D.; Chen, P.; Zeng, R.; Du, Q.; Tan, M.; and Gan, C. 2020.
\newblock {Location-aware Graph Convolutional Networks for Video Question
  Answering}.
\newblock In \emph{AAAI}.

\bibitem[{Jang et~al.(2017)Jang, Song, Yu, Kim, and Kim}]{jang2017}
Jang, Y.; Song, Y.; Yu, Y.; Kim, Y.; and Kim, G. 2017.
\newblock TGIF-QA: Toward Spatio-Temporal Reasoning in Visual Question
  Answering.
\newblock In \emph{CVPR}.

\bibitem[{Jhuang et~al.(2011)Jhuang, Garrote, Poggio, Serre, and
  Hmdb}]{jhuang2011large}
Jhuang, H.; Garrote, H.; Poggio, E.; Serre, T.; and Hmdb, T. 2011.
\newblock A large video database for human motion recognition.
\newblock In \emph{Proc. of IEEE International Conference on Computer Vision},
  volume~4, 6.

\bibitem[{Jiang and Han(2020)}]{jiang2020}
Jiang, P.; and Han, Y. 2020.
\newblock Reasoning with Heterogeneous Graph Alignment for Video Question
  Answering.
\newblock In \emph{Proceedings of the AAAI Conference on Artificial
  Intelligence}.

\bibitem[{Karpathy et~al.(2014)Karpathy, Toderici, Shetty, Leung, Sukthankar,
  and Fei-Fei}]{karpathy2014large}
Karpathy, A.; Toderici, G.; Shetty, S.; Leung, T.; Sukthankar, R.; and Fei-Fei,
  L. 2014.
\newblock Large-scale video classification with convolutional neural networks.
\newblock In \emph{Proceedings of the IEEE conference on Computer Vision and
  Pattern Recognition}, 1725--1732.

\bibitem[{Kay et~al.(2017)Kay, Carreira, Simonyan, Zhang, Hillier,
  Vijayanarasimhan, Viola, Green, Back, Natsev et~al.}]{kay2017kinetics}
Kay, W.; Carreira, J.; Simonyan, K.; Zhang, B.; Hillier, C.; Vijayanarasimhan,
  S.; Viola, F.; Green, T.; Back, T.; Natsev, P.; et~al. 2017.
\newblock The kinetics human action video dataset.
\newblock \emph{arXiv preprint arXiv:1705.06950} .

\bibitem[{Kim et~al.(2017)Kim, Heo, Choi, and Zhang}]{kim2017}
Kim, K.~M.; Heo, M.~O.; Choi, S.~H.; and Zhang, B.~T. 2017.
\newblock {Deepstory: Video story QA by deep embedded memory networks}.
\newblock \emph{IJCAI International Joint Conference on Artificial
  Intelligence} 2016--2022.
\newblock ISSN 10450823.

\bibitem[{Kingma and Ba(2015)}]{KingmaB14}
Kingma, D.~P.; and Ba, J. 2015.
\newblock Adam: {A} Method for Stochastic Optimization.
\newblock In Bengio, Y.; and LeCun, Y., eds., \emph{3rd International
  Conference on Learning Representations, {ICLR} 2015, San Diego, CA, USA, May
  7-9, 2015, Conference Track Proceedings}.
\newblock \urlprefix\url{http://arxiv.org/abs/1412.6980}.

\bibitem[{Ko{\v{c}}isk{\`y} et~al.(2018)Ko{\v{c}}isk{\`y}, Schwarz, Blunsom,
  Dyer, Hermann, Melis, and Grefenstette}]{kovcisky2018narrativeqa}
Ko{\v{c}}isk{\`y}, T.; Schwarz, J.; Blunsom, P.; Dyer, C.; Hermann, K.~M.;
  Melis, G.; and Grefenstette, E. 2018.
\newblock The narrativeqa reading comprehension challenge.
\newblock \emph{Transactions of the Association for Computational Linguistics}
  6: 317--328.

\bibitem[{Krishna et~al.(2017)Krishna, Hata, Ren, Fei-Fei, and
  Carlos~Niebles}]{krishna2017dense}
Krishna, R.; Hata, K.; Ren, F.; Fei-Fei, L.; and Carlos~Niebles, J. 2017.
\newblock Dense-captioning events in videos.
\newblock In \emph{Proceedings of the IEEE international conference on computer
  vision}, 706--715.

\bibitem[{Lei et~al.(2018)Lei, Yu, Bansal, and Berg}]{lei2018tvqa}
Lei, J.; Yu, L.; Bansal, M.; and Berg, T.~L. 2018.
\newblock TVQA: Localized, Compositional Video Question Answering.
\newblock In \emph{EMNLP}.

\bibitem[{Lei et~al.(2019)Lei, Yu, Berg, and Bansal}]{lei2019tvqa+}
Lei, J.; Yu, L.; Berg, T.~L.; and Bansal, M. 2019.
\newblock {TVQA+:} Spatio-Temporal Grounding for Video Question Answering.
\newblock \emph{CoRR} abs/1904.11574.
\newblock \urlprefix\url{http://arxiv.org/abs/1904.11574}.

\bibitem[{Li et~al.(2019)Li, Song, Gao, Liu, Huang, Gan, and He}]{li2019}
Li, X.; Song, J.; Gao, L.; Liu, X.; Huang, W.; Gan, C.; and He, X. 2019.
\newblock Beyond RNNs: Positional Self-Attention with Co-Attention for Video
  Question Answering.
\newblock \emph{AAAI} 8658--8665.

\bibitem[{Maharaj et~al.(2017)Maharaj, Ballas, Rohrbach, Courville, and
  Pal}]{maharaj2017}
Maharaj, T.; Ballas, N.; Rohrbach, A.; Courville, A.; and Pal, C. 2017.
\newblock A Dataset and Exploration of Models for Understanding Video Data
  Through Fill-In-The-Blank Question-Answering.
\newblock In \emph{The IEEE Conference on Computer Vision and Pattern
  Recognition (CVPR)}.

\bibitem[{Mclaughlin(1963)}]{mclaughlin1963psycho}
Mclaughlin, G.~H. 1963.
\newblock Psycho-logic: A possible alternative to Piaget's formulation.
\newblock \emph{British Journal of Educational Psychology} 33(1): 61--67.

\bibitem[{Mostafazadeh et~al.(2016)Mostafazadeh, Chambers, He, Parikh, Batra,
  Vanderwende, Kohli, and Allen}]{mostafazadeh2016corpus}
Mostafazadeh, N.; Chambers, N.; He, X.; Parikh, D.; Batra, D.; Vanderwende, L.;
  Kohli, P.; and Allen, J. 2016.
\newblock A corpus and cloze evaluation for deeper understanding of commonsense
  stories.
\newblock In \emph{Proceedings of the 2016 Conference of the North American
  Chapter of the Association for Computational Linguistics: Human Language
  Technologies}, 839--849.

\bibitem[{Mueller(2004)}]{mueller2004understanding}
Mueller, E.~T. 2004.
\newblock Understanding script-based stories using commonsense reasoning.
\newblock \emph{Cognitive Systems Research} 5(4): 307--340.

\bibitem[{Mun et~al.(2017)Mun, Seo, Jung, and Han}]{mun2017marioQA}
Mun, J.; Seo, P.~H.; Jung, I.; and Han, B. 2017.
\newblock MarioQA: Answering Questions by Watching Gameplay Videos.
\newblock In \emph{ICCV}.

\bibitem[{Pascual-Leone(1969)}]{pascual1969cognitive}
Pascual-Leone, J. 1969.
\newblock \emph{Cognitive development and cognitive style: A general
  psychological integration}.

\bibitem[{Pennington, Socher, and Manning(2014)}]{pennington2014glove}
Pennington, J.; Socher, R.; and Manning, C.~D. 2014.
\newblock GloVe: Global Vectors for Word Representation.
\newblock In \emph{Empirical Methods in Natural Language Processing (EMNLP)},
  1532--1543.
\newblock \urlprefix\url{http://www.aclweb.org/anthology/D14-1162}.

\bibitem[{Perazzi et~al.(2016)Perazzi, Pont-Tuset, McWilliams, Van~Gool, Gross,
  and Sorkine-Hornung}]{perazzi2016benchmark}
Perazzi, F.; Pont-Tuset, J.; McWilliams, B.; Van~Gool, L.; Gross, M.; and
  Sorkine-Hornung, A. 2016.
\newblock A benchmark dataset and evaluation methodology for video object
  segmentation.
\newblock In \emph{Proceedings of the IEEE Conference on Computer Vision and
  Pattern Recognition}, 724--732.

\bibitem[{Piaget(1972)}]{piaget1972intellectual}
Piaget, J. 1972.
\newblock Intellectual evolution from adolescence to adulthood.
\newblock \emph{Human development} 15(1): 1--12.

\bibitem[{Richardson, Burges, and Renshaw(2013)}]{richardson2013mctest}
Richardson, M.; Burges, C.~J.; and Renshaw, E. 2013.
\newblock Mctest: A challenge dataset for the open-domain machine comprehension
  of text.
\newblock In \emph{Proceedings of the 2013 Conference on Empirical Methods in
  Natural Language Processing}, 193--203.

\bibitem[{Riedl(2016)}]{riedl2016computational}
Riedl, M.~O. 2016.
\newblock Computational narrative intelligence: A human-centered goal for
  artificial intelligence.
\newblock \emph{arXiv preprint arXiv:1602.06484} .

\bibitem[{Riedl and Young(2010)}]{riedl2010narrative}
Riedl, M.~O.; and Young, R.~M. 2010.
\newblock Narrative planning: Balancing plot and character.
\newblock \emph{Journal of Artificial Intelligence Research} 39: 217--268.

\bibitem[{Rohrbach et~al.(2017)Rohrbach, Torabi, Rohrbach, Tandon, Pal,
  Larochelle, Courville, and Schiele}]{rohrbach2017movie}
Rohrbach, A.; Torabi, A.; Rohrbach, M.; Tandon, N.; Pal, C.; Larochelle, H.;
  Courville, A.; and Schiele, B. 2017.
\newblock Movie description.
\newblock \emph{International Journal of Computer Vision} 123(1): 94--120.

\bibitem[{Schank and Abelson(2013)}]{schank2013scripts}
Schank, R.~C.; and Abelson, R.~P. 2013.
\newblock \emph{Scripts, plans, goals, and understanding: An inquiry into human
  knowledge structures}.
\newblock Psychology Press.

\bibitem[{Seo et~al.(2016)Seo, Kembhavi, Farhadi, and
  Hajishirzi}]{Seo2016BidirectionalAF}
Seo, M.; Kembhavi, A.; Farhadi, A.; and Hajishirzi, H. 2016.
\newblock Bidirectional Attention Flow for Machine Comprehension.
\newblock \emph{ArXiv} abs/1611.01603.

\bibitem[{Seol, Sharp, and Kim(2011)}]{seol2011stanford}
Seol, S.; Sharp, A.; and Kim, P. 2011.
\newblock Stanford Mobile Inquiry-based Learning Environment (SMILE): using
  mobile phones to promote student inquires in the elementary classroom.
\newblock In \emph{Proceedings of the International Conference on Frontiers in
  Education: Computer Science and Computer Engineering (FECS)}, 1.

\bibitem[{Soomro, Zamir, and Shah(2012)}]{soomro2012dataset}
Soomro, K.; Zamir, A.~R.; and Shah, M. 2012.
\newblock A dataset of 101 human action classes from videos in the wild.
\newblock \emph{Center for Research in Computer Vision} 2.

\bibitem[{Sukhbaatar et~al.(2015)Sukhbaatar, szlam, Weston, and
  Fergus}]{sukhbaatar2015}
Sukhbaatar, S.; szlam, a.; Weston, J.; and Fergus, R. 2015.
\newblock End-To-End Memory Networks.
\newblock In Cortes, C.; Lawrence, N.~D.; Lee, D.~D.; Sugiyama, M.; and
  Garnett, R., eds., \emph{Advances in Neural Information Processing Systems
  28}, 2440--2448. Curran Associates, Inc.
\newblock
  \urlprefix\url{http://papers.nips.cc/paper/5846-end-to-end-memory-networks.pdf}.

\bibitem[{Szilas(1999)}]{szilas1999interactive}
Szilas, N. 1999.
\newblock Interactive drama on computer: beyond linear narrative.
\newblock In \emph{Proceedings of the AAAI fall symposium on narrative
  intelligence}, 150--156.

\bibitem[{Tapaswi et~al.(2016)Tapaswi, Zhu, Stiefelhagen, Torralba, Urtasun,
  and Fidler}]{tapaswi2015}
Tapaswi, M.; Zhu, Y.; Stiefelhagen, R.; Torralba, A.; Urtasun, R.; and Fidler,
  S. 2016.
\newblock {MovieQA: Understanding Stories in Movies through
  Question-Answering}.
\newblock In \emph{IEEE Conference on Computer Vision and Pattern Recognition
  (CVPR)}.

\bibitem[{Trischler et~al.(2016)Trischler, Wang, Yuan, Harris, Sordoni,
  Bachman, and Suleman}]{trischler2016newsqa}
Trischler, A.; Wang, T.; Yuan, X.; Harris, J.; Sordoni, A.; Bachman, P.; and
  Suleman, K. 2016.
\newblock Newsqa: A machine comprehension dataset.
\newblock \emph{arXiv preprint arXiv:1611.09830} .

\bibitem[{Winston(2011)}]{winston2011strong}
Winston, P.~H. 2011.
\newblock The strong story hypothesis and the directed perception hypothesis.
\newblock In \emph{2011 AAAI Fall Symposium Series}.

\bibitem[{Wolf et~al.(2020)Wolf, Debut, Sanh, Chaumond, Delangue, Moi, Cistac,
  Rault, Louf, Funtowicz, Davison, Shleifer, von Platen, Ma, Jernite, Plu, Xu,
  Scao, Gugger, Drame, Lhoest, and Rush}]{wolf-etal-2020-transformers}
Wolf, T.; Debut, L.; Sanh, V.; Chaumond, J.; Delangue, C.; Moi, A.; Cistac, P.;
  Rault, T.; Louf, R.; Funtowicz, M.; Davison, J.; Shleifer, S.; von Platen,
  P.; Ma, C.; Jernite, Y.; Plu, J.; Xu, C.; Scao, T.~L.; Gugger, S.; Drame, M.;
  Lhoest, Q.; and Rush, A.~M. 2020.
\newblock Transformers: State-of-the-Art Natural Language Processing.
\newblock In \emph{Proceedings of the 2020 Conference on Empirical Methods in
  Natural Language Processing: System Demonstrations}, 38--45. Online:
  Association for Computational Linguistics.
\newblock \urlprefix\url{https://www.aclweb.org/anthology/2020.emnlp-demos.6}.

\bibitem[{Xu et~al.(2016)Xu, Mei, Yao, and Rui}]{xu2016msr}
Xu, J.; Mei, T.; Yao, T.; and Rui, Y. 2016.
\newblock Msr-vtt: A large video description dataset for bridging video and
  language.
\newblock In \emph{Proceedings of the IEEE conference on computer vision and
  pattern recognition}, 5288--5296.

\bibitem[{Yi et~al.(2020)Yi, Gan, Li, Kohli, Wu, Torralba, and
  Tenenbaum}]{yi2020clevrer}
Yi, K.; Gan, C.; Li, Y.; Kohli, P.; Wu, J.; Torralba, A.; and Tenenbaum, J.~B.
  2020.
\newblock {CLEVRER:} Collision Events for Video Representation and Reasoning.
\newblock In \emph{ICLR}.

\end{thebibliography}

\clearpage
\pagebreak

\appendix
\addcontentsline{toc}{section}{Appendices}
\renewcommand{\thesection}{\Alph{section}.\arabic{section}}
\begin{appendix}

\begin{center}
    \textbf{\LARGE APPENDIX}
\end{center}
This appendix provides additional information not described in the main text due to the page limit. It contains additional related works in Section A, analyses about the dataset in Section B.1, dataset collection methods in Section B.2, implementation details in Section C, comparative experimental results of our model and other methods in Section D, and qualitative results of our model in Section E.

\begin{table*}[h]
 \begin{tabular}{c @{\hskip 0.2in}c @{\hskip 0.2in} c @{\hskip 0.2in}c @{\hskip 0.2in}c@{\hskip 0.2in}c@{\hskip 0.2in}c@{\hskip 0.2in}c@{\hskip 0.2in}c} 
 \specialrule{.1em}{.05em}{.05em} 
     & \multirow{2}{*}{\# QAs}
     & \multirow{2}{*}{\# Clips}
     & \multirow{2}{*}{\begin{tabular}{@{}c@{}} Avg. Video Len \\ Shot / Scene \end{tabular}}
     & \multirow{2}{*}{\# Annotated Images}  
     & \multicolumn{4}{c}{\# QAs by Difficulty}\\
     \cline{6-9}
     &&&&& 1 &2 &3 &4 \\
 \specialrule{.1em}{.05em}{.05em} 
    Train    & {\trainqa}         & 8,976    & 3.5 / 93.0     & 130,924 & 5,799& 2,826& 1,249& 1,244\\
    Val       &  {\valqa}     & 2,691     & 3.5 / 87.0     & 41,049 & 1,732& 851& 416& 413 \\
    Test        &  {\testqa}         & 2,767     & 3.8 / 93.0    & 45,033&1,782& 853& 409 & 409 \\
    \hline
    Total       &  {\qa}       & 14,434     & 3.6 / 91.8     & 217,006  & 9,313& 4,530& 2,074& 2,066\\
 \specialrule{.1em}{.05em}{.05em} 
 \end{tabular}
 \centering
 \caption{Statistics about train, validation, and test split of DramaQA dataset. \# QAs: The number of QA pairs. \# Clips: The number of video clips including shot and scene. Avg. Video Len: Average video length per each video clip. \# Annotated Images: The number of annotated images in total target video. \# QAs by Difficulty: The number of QA pairs for each difficulty level.}
 \label{table:stats_split}
 \end{table*}

\section{A. Additional Related Work}
\label{app:relatedwork}
\subsection{Video Understanding} 
Video understanding area has been actively studied and several tasks including datasets have been proposed such as action recognition~\cite{jhuang2011large,soomro2012dataset,karpathy2014large,caba2015activitynet,kay2017kinetics}, video classification~\cite{abu2016youtube}, video captioning~\cite{xu2016msr,krishna2017dense}, and spatio-temporal object segmentation~\cite{perazzi2016benchmark}. However, these researches focus on perceiving and recognizing visual elements so that they are not suitable for high-level visual reasoning.

To circumvent this limitation, video question answering for short video clips is proposed as a benchmark for high-level video understanding~\cite{rohrbach2017movie,jang2017}, which is followed by many works utilizing spatio-temporal reasoning. \cite{gao2019} and \cite{li2019} used the attention mechanism as a reasoning module, and \cite{jiang2020} and \cite{huang2020} utilized graphs for modeling high-level temporal structure. Also, several datasets are proposed for temporal reasoning~\cite{yi2020clevrer, girdhar2020cater}, which shows intensive attraction to reasoning over the temporal structure. But, these works only dealt with a sequence of images about short video clips, which would not include a meaningful story.

Unlike these video question answering, Video Story Question Answering focuses on the story narrative about the video. A story is a sequence of events, which means video with meaningful stories contains a relatively long sequence of videos and consists of a series of correlated video events. Video Story Question Answering requires the ability to discriminate what's meaningful in a very long video and also requires visual processing, natural language, and additional acoustic modeling. 

\section{B. Dataset}
\label{app:dataset}
\subsection{B.1 Dataset Analysis}
\label{app:analysis}
\subsubsection{Data Split}
DramQA dataset has {\qa} QA pairs, and we separated them into train/val/test split by 60\%/20\%/20\%, respectively, according to the non-overlapping videos so that they do not refer to the same video. We make sure that the videos used for training are not used for validation and test, which is important to evaluate the performance.
Table \ref{table:stats_split} shows statistics about DramaQA dataset.

\subsubsection{Hierarchical QAs}
\begin{figure}[!h]
 \centering
 \includegraphics[width=1.0\columnwidth]{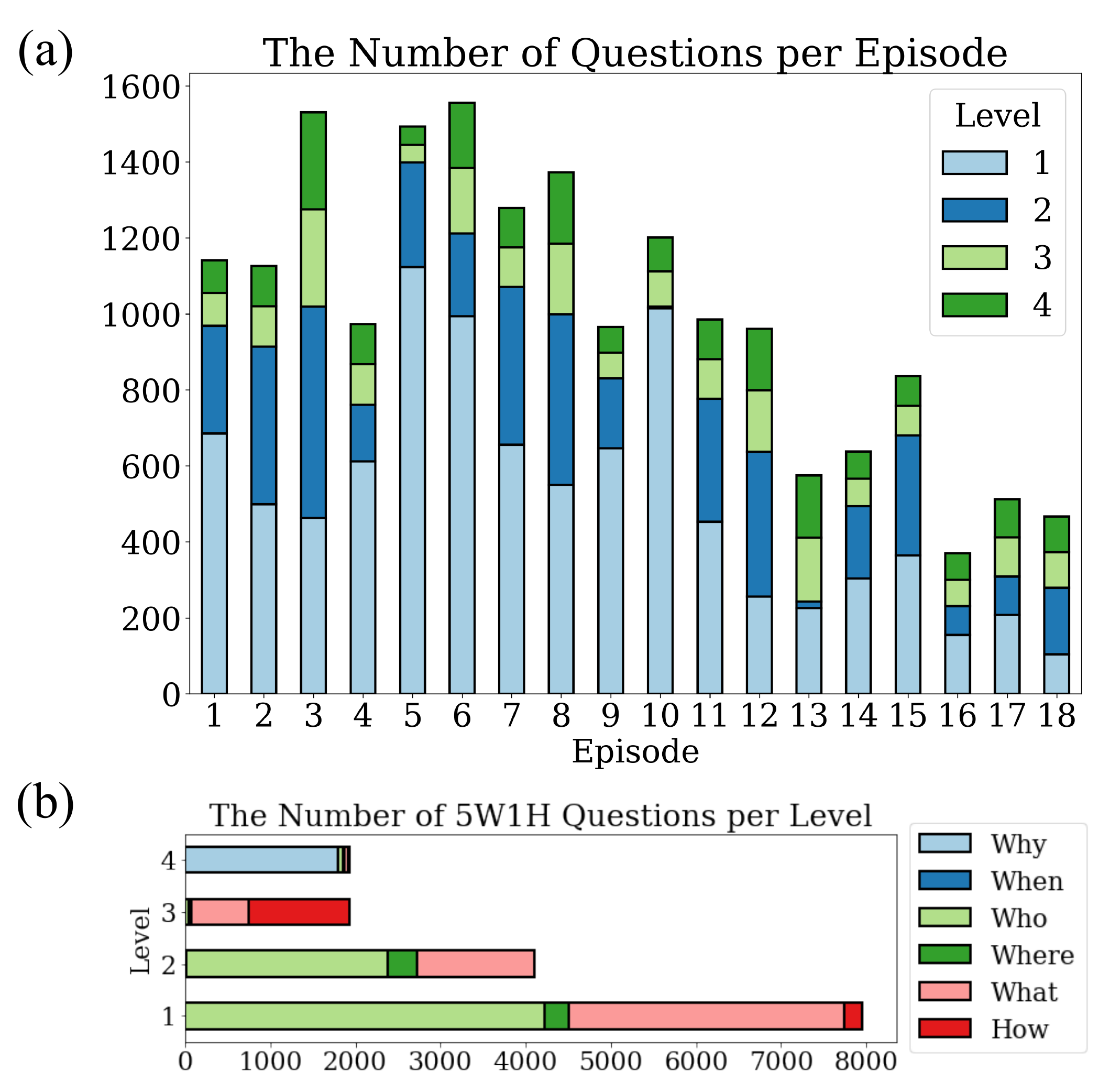}
 \caption{\textbf{(a)} The number of QA pairs per episode and difficulty level. Given that the length of scene is tens of times longer than the size of shot, the variation between levels is small compared to the number of videos. \textbf{(b)} The number of 5W1H question types per difficulty level.}
 \label{fig:stats_question}
 \end{figure}

DramaQA has four hierarchical difficulty levels for each question, which are used as an evaluation metric for video story understanding. Figure~\ref{fig:stats_question}(a) shows overall distributions of levels along with the episodes. Since a single \textit{scene} of a video has multiple \textit{shots}, the number of questions for Difficulty 1 and 2 is naturally larger than that for Difficulty 3 and 4.
We also visualize 5W1H question types per difficulty level in Figure~\ref{fig:stats_question}(b).
In Difficulty 1 and 2, \texttt{Who} and \texttt{What} questions are the majority. In case of Difficulty 3, \texttt{How} and \texttt{What} types are the top-2 questions. In Difficulty 4, most of questions start with \texttt{Why}. 

 \index{queperepiandlevel}

\subsubsection{Visual Metadata}
To collect character-centered visual metadata, we predefined main characters, emotions, and behaviors as follows:
 \begin{itemize}
    \item Main character: Anna, Chairman, Deogi, Dokyung, Gitae, Haeyoung1, Haeyoung2, Heeran, Hun, Jeongsuk, Jinsang, Jiya, Kyungsu, Sangseok, Seohee, Soontack, Sukyung, Sungjin, Taejin, Yijoon
     \item Emotion: anger, disgust, fear, happiness, sadness, surprise, neutral
     \item Behavior: drink, hold, point out, put arms around each other's shoulder, clean, cook, cut, dance, destroy, eat, look for, high-five, hug, kiss, look at/back on, nod, open, call, play instruments, push away, shake hands, sing, sit down, smoke, stand up, walk, watch, wave hands, write
 \end{itemize}
 
\begin{figure*}[!h]
 \centering
 \includegraphics[width=0.9\textwidth]{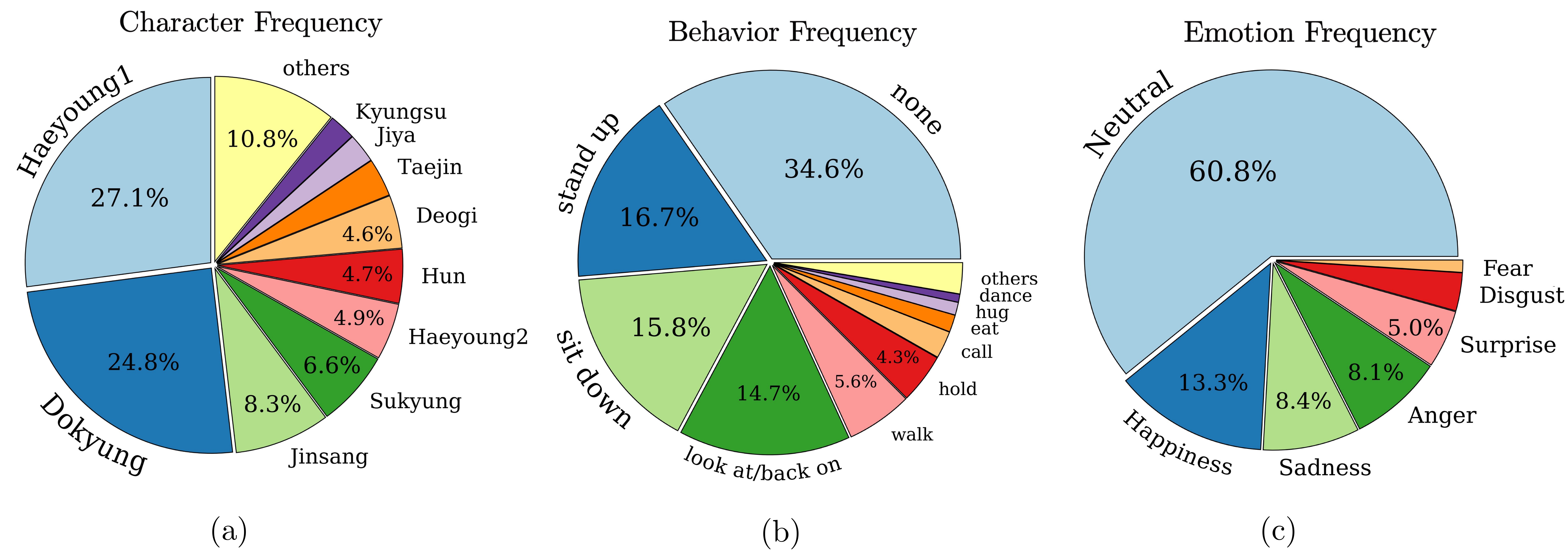}
 \caption{\textbf{(a)} The percentage of each character's frequency in visual metadata. \textit{Haeyoung1} and \textit{Dokyung} are two main characters of drama \textit{AnotherMissOh}. \textit{Haeyoung2} is the person who has same name with \textit{Haeyoung1}, but we divided their name with numbers to get rid of confusion. \textbf{(b)} The percentage of each behavior frequency in the visual metadata. \texttt{none} behavior occupies a lot because there are many frames with only character's face. \textbf{(c)} The percentage of each emotion frequency in the visual metadata.}
 \label{fig:stats2}
 \end{figure*}

In Figure~\ref{fig:stats2}, distributions of main character and their behavior and emotion in visual metadata are illustrated. As shown in Figure~\ref{fig:stats2}(a), \textit{Haeyoung1} and \textit{Dokyung} appear the most frequently among all characters. Due to the nature of the TV drama, it is natural that the frequency of appearance varies depending on the importance of the characters: long-tail distribution. For Figure~\ref{fig:stats2}(b) and (c), various behaviors and emotions are appeared except the situations when behavior and emotion cannot express much information due to their own trait like none behavior and neutral emotion.  Also, it is natural for behavior and emotion to follow long-tail distribution as reported in MovieGraph~\cite{vicol2018}, and TVQA+~\cite{lei2019tvqa+}.

\subsubsection{Coreference Resolved Scripts}
To understand video stories, especially drama, it is crucial to understand the dialogue between the characters.
Notably, the information such as ``\textit{Who} is talking to \textit{whom} about \textit{who} did what?'' is significant in order to understand whole stories.

\begin{figure*}[!h]
 \centering
 \includegraphics[width=0.8\textwidth]{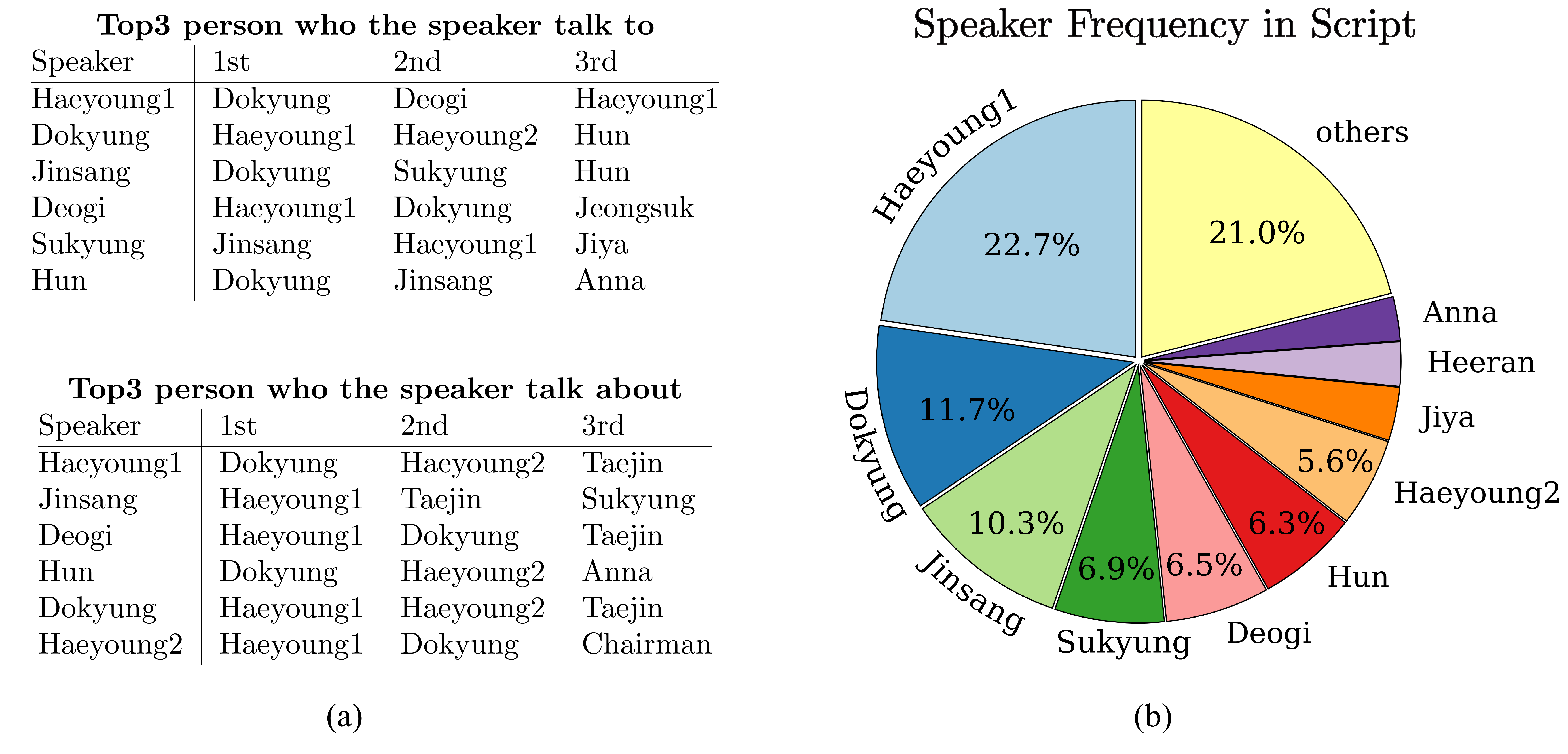}
 \caption{\textbf{(a)} \textbf{Top:} Top-3 the person who the speaker frequently talks to, for each top 6 most spoken person. \textbf{Bottom:} Top-3 the person who the speaker frequently talks about, for each top 6 most spoken person. \textbf{(b)} The percentage of each person's utterance in the script.}
 \label{fig:stats1}
 \end{figure*}

In Figure \ref{fig:stats1}, we show analyses of each person's utterances in the scripts. First of all, we analyze who talks the most with the main character and who is mentioned in their dialogue. As shown in Figure \ref{fig:stats1}(a), we can see that two protagonists of the drama, \textit{Dokyung} and \textit{Haeyoung1}, appeared most often in their dialogue. Also, it indirectly shows the relationship between the main characters. \textit{Hun}, \textit{Dokyung}'s brother, is familiar to \textit{Dokyung} but a stranger to \textit{Haeyoung1}. Figure \ref{fig:stats1}(b) shows the percentage of each character's utterance from whole episodes.

 \index{qaperepi}

\subsection{B.2 Data Collection}
\label{app:collection}
 Since annotating information for video drama requires not only knowledge of the characters but also the stories going on the drama, utilizing a crowdsourcing service might decrease the quality of the resulting annotation dataset.
 Therefore, with semi-automated visual annotation tagging tools, all annotation tasks were carried out by a small group of dedicated workers (20 workers) who are aware of the whole drama story-line and the individual characters involved in the story 
 
 \subsubsection{Worker Verification}
  Before starting to collect data, we held one day training session to fully understand the annotation guidelines. During the data collecting process, we worked closely with them to obtain highly reliable annotations. The intermediate annotation results were checked and feedback was given to continue the work. Hired workers didn’t start to make their own QAs until they fully understood our hierarchical difficulties. After the annotation procedure, data inspection was proceeded to confirm their QAs were made correctly.
 
 \subsubsection{Video Separation}
 For video separation, every shot boundary in each episode is pre-detected automatically with the sum of the absolute difference between successive frames' color. As a scene is multiple shots that take place in the same location with semantic connections between themselves, we didn't use automatic tools for making a scene. Rather, we manually merged successive shots that have the same location to build scene-level video clips. 

 
 \subsubsection{Visual Metatdata Annotation}
 For visual metadata annotation, the visual bounding boxes were created using an object detection and human/face detection tool (YOLOv3), and workers manually fine-tuned the bounding boxes and the main characters' names, behaviors, and emotions. 
 
 \subsubsection{Collection Guidelines for QA}
 The QA pairs were created with the following rules: 1) Workers must use the main characters' names (i.e. \textit{Haeyoung1}, \textit{Dokyung}, \textit{Deogi}, and etc.) instead of pronouns (i.e. \textit{They}, \textit{He}, \textit{She}, etc.). 2) Questions and answers should be in complete sentences. 3) All sentences should be case-sensitive. For hierarchical difficulty of QAs, different rules were applied for each level:
 \begin{itemize}
     \item Difficulty 1
     \begin{itemize}
         \item The Question-Answering (QA) set should be based on only one supporting fact (\textit{A triplet form of  subject-relationship-object})  from a video. 
         \item Question can (should) be started with \textit{Who}, \textit{Where}, and \textit{What}.
     \end{itemize}
     \item Difficulty 2
     \begin{itemize}
         \item The Question-Answering (QA) set should be based on multiple supporting facts from a video. 
         \item Question can (should) be started with \textit{Who}, \textit{Where}, and \textit{What}.
     \end{itemize}
     \item Difficulty 3
     \begin{itemize}
         \item The Question-Answering (QA) set should be based on multiple situations/actions with sequential information. To answer the question at this difficulty, temporal connection of multiple supporting facts should be considered, differently from Difficulty 2 questions.
         \item Question can be started with \textit{How} (recommended) and \textit{What}.
     \end{itemize}
     \item Difficulty 4
     \begin{itemize}
         \item The Question-Answering (QA) set should be based on reasoning for causality.
         \item Question can be started with \textit{Why}.\footnote{We clarified it to restrict causality in QA procedure and also it’s natural to have “why question”, due to the nature of the difficulty of inferring causality.}
     \end{itemize}
 \end{itemize}
 
 \section{C. Implementation Details}
 \label{app:imp_details}
 
 We used pretrained GloVe features~\cite{pennington2014glove} to initialize and make each word embedding trainable. We also initialized the main character words with random vectors to represent character level features. For visual bounding boxes, we used pretrained ResNet-18~\cite{He2015DeepRL} to extract features. To get streams with temporal context from input streams, we used Bidirectional LSTM with 300 hidden dimensions. 
 We limited our model to use less than 30 sentences for script, 30 shots for visual inputs, words lengths per sentence to 20 words, and frames per shot to 10 frames for memory usage. The batch size is 4 and cross-entropy loss is used. We used Adam~\cite{KingmaB14} with $10^{-4}$ learning rate and $10^{-5}$ weight decay as a regularization.
\begin{table*}[h!]

\begin{tabular}{l @{\hskip 0.2in} c @{\hskip 0.2in} c @{\hskip 0.2in}c @{\hskip 0.2in} c @{\hskip 0.1in} | @{\hskip 0.1in} c @{\hskip 0.1in} c @{\hskip 0.1in} c } 
\specialrule{.1em}{.05em}{.05em} 
    Model
     & Diff. 1
     & Diff. 2
     & Diff. 3 
     & Diff. 4
     & Overall 
     & Diff. Avg.\\
\specialrule{.1em}{.05em}{.05em} 
    
    Shortest Answer     & 20.54  & 18.64  & 20.78  & 18.83  & 19.90  & 19.70  \\
    Longest Answer    & 23.85   & 21.10  & 31.05  & 30.81  & 24.85  & 26.70  \\
    QA Similarity & 30.64   & 27.20   & 26.16   & 22.25    & 28.27   & 26.56  \\
    QA+V+S & 57.24&	49.12 & 41.32 & 39.85 & 51.29 & 46.88 \\
    QA+BERT & 53.93 & 50.76 &	49.14&	51.83&	52.33 & 51.42\\
    \hline
    MemN2N~\cite{tapaswi2015} & 59.82 & 52.64 & 41.56 & 38.63 & 53.37 & 48.16 \\
    MemN2N$+$annotation & 64.39&58.15&44.74&42.79&57.96&53.13\\
    TVQA~\cite{lei2018tvqa} & 66.05&	61.78&	53.79&	53.55&	62.06& 58.79  \\
    TVQA$+$annotaion &	74.80&	72.57&	53.30&	55.75&	69.45&64.10\\
    \hline
    Our$-$annotation  & 65.99&65.30&55.99&58.68&63.77&61.49\\
    Our (Full)  & 75.96  & 74.65  & 57.36  & 56.63  & 71.14   & 66.15   \\
\specialrule{.1em}{.05em}{.05em} 
\end{tabular}
\centering
\caption{Quantitative result for our model on test split. Last two columns show the performance of overall test split and the average performance of each set.  \texttt{Shortest}, \texttt{Longest}, \texttt{QA Similarity}, and \texttt{QA+V+S} are baseline methods. \texttt{MemN2N} and \texttt{TVQA} are proposed baseline methods in other datasets. \texttt{$-$annotation} removed coreference and visual metadata from inputs. Details of each model are described in the Appendix D.}
\label{table:baseline}
\end{table*}

 \section{D. Additional Quantitative Results}
 \label{app:quantitative}
 For more detailed analysis of the proposed model and dataset, we report the comparative results with previously proposed methods for other video QA datasets as well as additional simple baselines.
 Since previous approaches were not designed to use additional annotations of DramaQA dataset, such as visual metadata, we also provide experimental results of modified models of them to utilize the additional annotations for fair comparison and evaluation of the effects of the proposed annotations.
 \subsection{D.1 Comparative Methods}
 
 
 \subsubsection{Shortest Answer:} This baseline selects the shortest answer for each question. 
 
 \subsubsection{Longest Answer: } This baseline selects the longest answer for each question. 
 
 \subsubsection{QA Similarity: } This baseline model is designed to choose the highest score on the cosine similarity between the average of question's word embeddings and the average of candidate answer's word embeddings. 

 \subsubsection{QA+V+S: } This baseline uses QA, visual inputs, and subtitles. We used the word embedding for textual inputs and visual features from images. Using bidirectional LSTM and mean pooling over time, we got temporal contexts from each input stream. Then, we concatenated the context representation for each input stream and got final score of each candidate answer with multilayer perceptrons. 
 
 \subsubsection{QA+BERT: } This baseline uses QA with BERT pretrained model~\cite{wolf-etal-2020-transformers}. We concatenate each QA pair and feed them into pretrained bert-base-cased model. Then we project the last hidden layer sequences down to a lower dimension (d = 300) with a single linear layer and took the max pooling across each sequence. Finally we computed a score for each answer using a single linear classifier.


 
 
 \subsubsection{MemN2N~\cite{tapaswi2015}}: MemN2N was proposed by \cite{tapaswi2015} which modified Memory Network~\cite{sukhbaatar2015} for multiple-choice video question and answering. The MemN2N utilizes (a set of) attention-based memory module to infer the answer from the question. 
 
 We also report revised version of MemN2N, \texttt{MemN2N$+$annotation} to use our additional annotations. We concatenated word embedding of visual metadata (name, behavior, and emotion) with visual bounding box features before the linear projection layer. Also, we concatenated the speaker and coreference in each sentence of scripts. 
 
 \subsubsection{TVQA~\cite{lei2018tvqa}}: TVQA model~\cite{lei2018tvqa} uses multiple streams for utilizing multimodal inputs. It encodes streams with multiple LSTM modules and get question-aware representations with  context matching modules to get final answer.
 We used the version of their \texttt{S+V(img)+Q} model for comparison. 

 Additionally, for the fair comparison, we changed visual concept features into visual metadata inputs of our dataset. Also, we used the speaker and coreference of our script as concatenated form and used bounding box information for extracting regional visual features. We used three input streams to get final score: subtitle, ImageNet features, and regional visual features. \texttt{TVQA$+$annotation} indicates our revised version of TVQA model. 

\begin{figure*}[!h]
 \centering
 \includegraphics[width=1.0\textwidth]{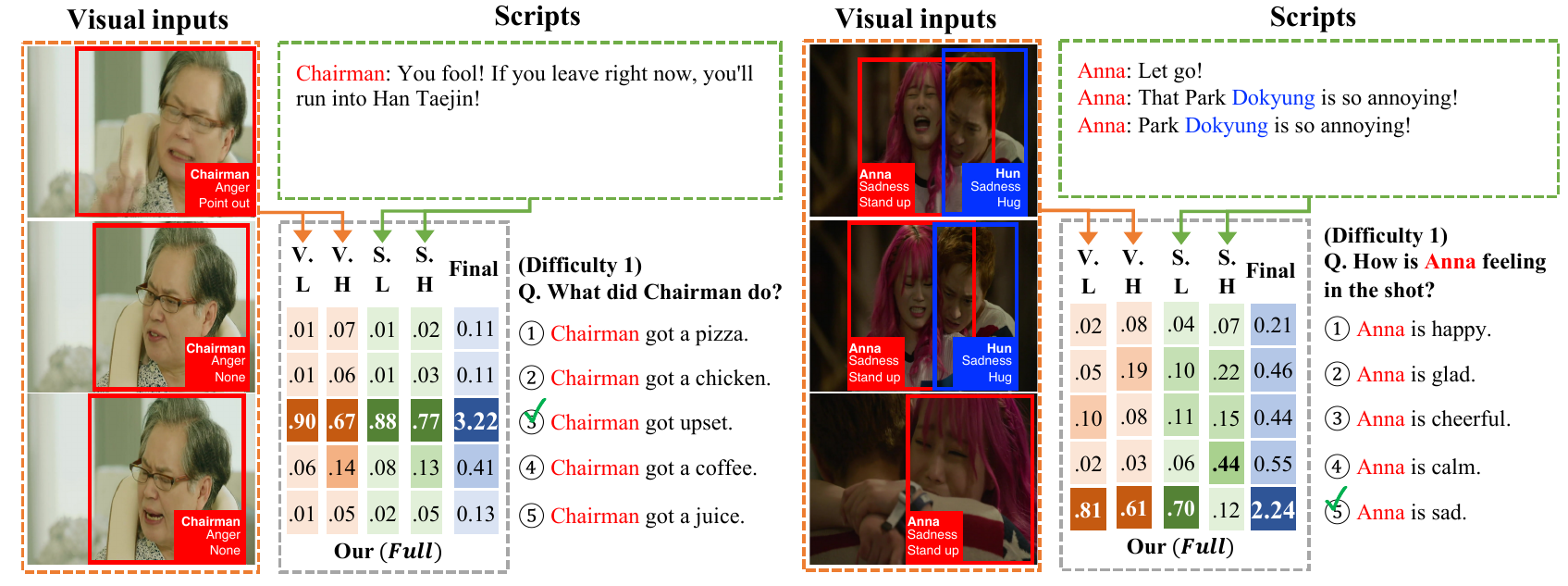}
 \caption{Examples of our model's correct prediction at Difficulty 1. We denote the scores from multi-level representations of each modality. \texttt{V.L} is the scores from the low-level representation of visual inputs, \texttt{V.H} is the scores from the high-level representation of visual inputs, \texttt{S.L} is the scores from the low-level representation of scripts, and \texttt{S.H} is the scores from the low-level representation of scripts. We denote the final scores in the blue colored table. A green checkmark indicates the model predictions, which is equal to the ground truth answer. }
 \label{fig:qual_result_app_low}
 \end{figure*}
 \index{figure}
 
\subsection{D.2 Results}
Table~\ref{table:baseline} shows the comparative results.
Firstly, we analyze more detailed characteristics with several baseline models.
As seen in line 1-2, reasoning the correct answers is hardly affected by the length of the them. 
However, in Diff.3 and 4, we can see that the correct answers tend to slightly longer than that of the wrong answers.  
Also, the results in line 3-4 showed that answering the questions with video contents is considerably accurate than without video contents.
From the result, we can confirm that the questions in our datasets are fairly related to the video contents.
Additionally, line 5 showed that there are superficial correlations between questions and answers. We guess this bias is due to the imbalanced occurrence of characters and objects, which is the nature of the TV drama.

Next, we compare our model with other Video QA models. By comparing line 6-7 and 8-9, we find that our character-centered annotations are always helpful irrespective of model architectures.
For the \texttt{MemN2N} (line 6-7), our approach has way better performance in overall.
For the \texttt{TVQA} (line 8-9), it shows good performance close to our model but still our model works better. Note that this model requires 16M parameters to train, while our model requires 7.7M parameters to train. Therefore, we can tell that the architectures of our model is more suitable for answer the question given video story. 

One thing we would like to note is that the architecture of the proposed model (Multi-level Character Matching) is designed by focusing more on reflecting various characteristics of our dataset rather than merely boosting the performance. Therefore, the performance could be improved by using more sophisticated modules, e.g., BERT~\cite{devlin2018pretraining}. We left those improvements to future research of other researchers.

\begin{figure*}[!h]
 \centering
 \includegraphics[width=1.0\textwidth]{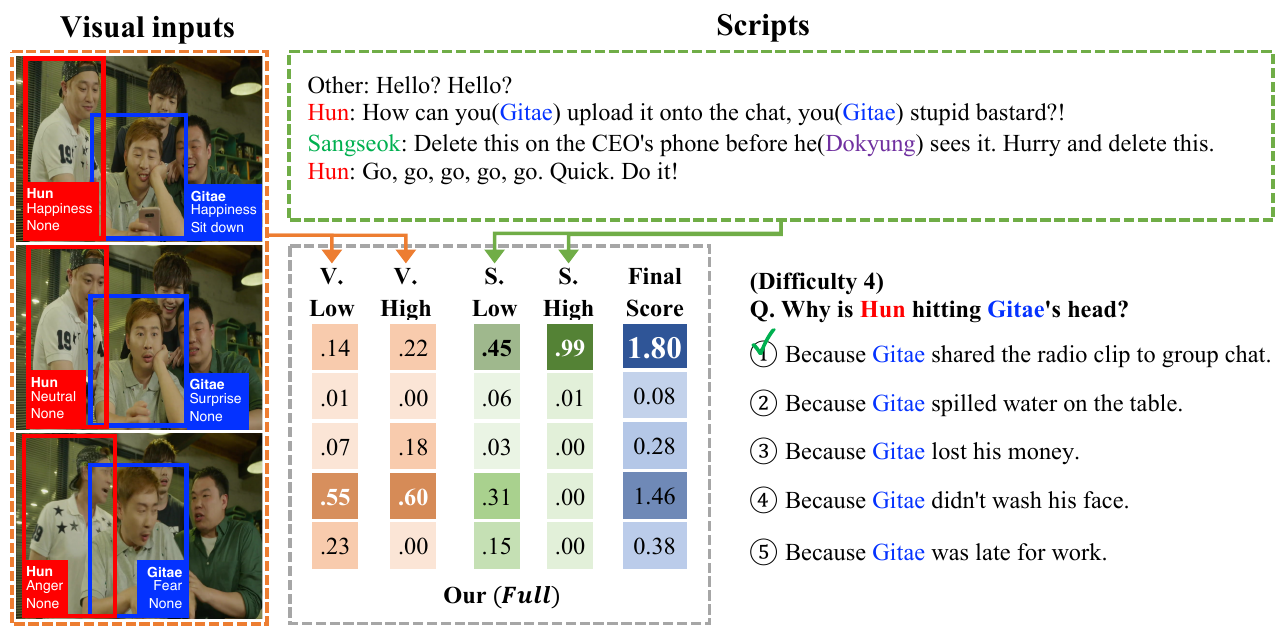}
 \caption{Examples of our model's correct prediction at Difficulty 4. We denote the scores from multi-level representations of each modality. A green checkmark indicates the model prediction which is equal to the ground truth answer. This case shows the model selects the best answer utilizing both modalities.}
 \label{fig:qual_result_app_correct}
 \end{figure*}
 \index{figure}

\begin{figure*}[!h]
 \centering
 \includegraphics[width=1.0\textwidth]{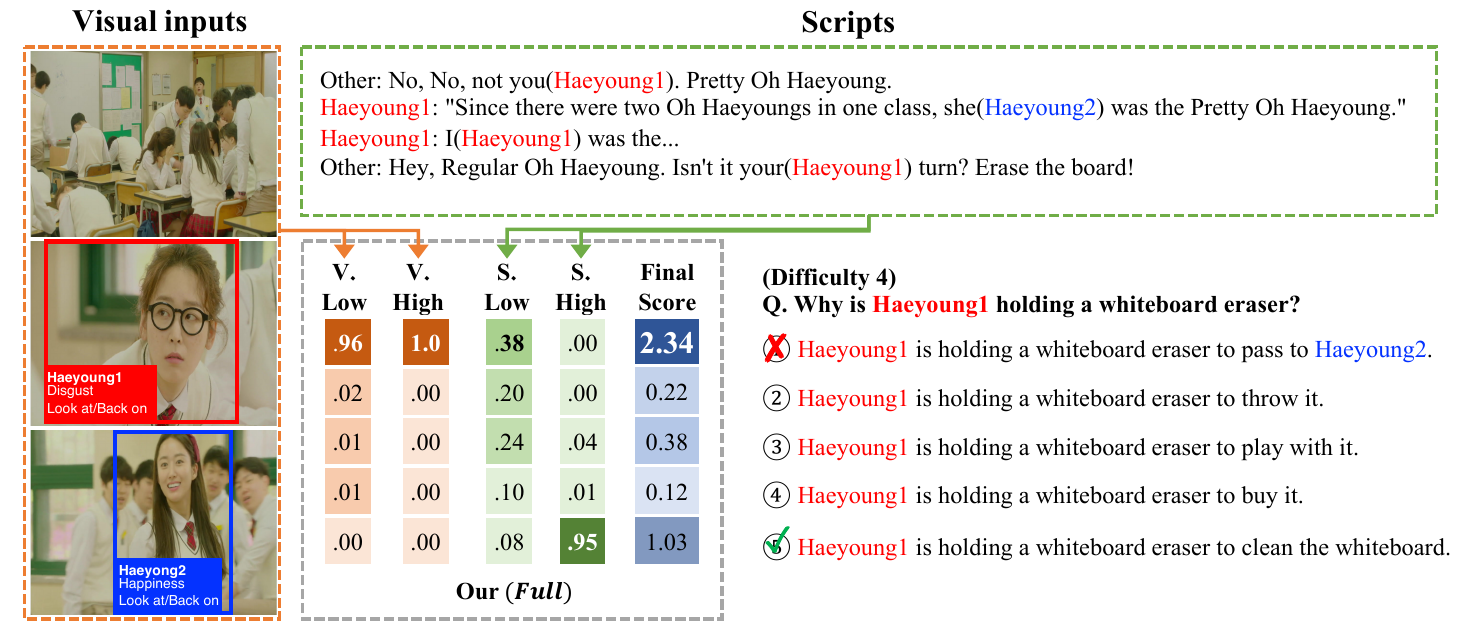}
 \caption{Examples of our model's failure to predict correct answer at Difficulty 4. We denote the scores from multi-level representations of each modality. Prediction of our model is indicated by red crossmark which is wrong answer, and the ground truth answer is indicated by green checkmark. This case shows the weakness of character-centered context matching.}
 \label{fig:qual_result_app_wrong}
 \vspace{-1.0em}
 \end{figure*}
 \index{figure}
 
 \section{E. Qualitative Results}
 \label{app:qualitative}
 Figure \ref{fig:qual_result_app_low} shows the examples of our model's prediction. 
 In the left example, we can find evidence for the correct answer from both textual and visual sources, so the scores from each source are both reasonable. 
 In the case of the right example, all but \texttt{S.H} predicted the correct answer. It's clear evidence to choose correct answer that visual metadata has the emotion \textit{Sadness} from \textit{Anna} and scripts have the word \textit{sad} from \textit{Hun}. However, in the scores from visual inputs, high-level representation has lower confidence than low-level representation, and in the scores from the script, high-level representation is choosing the wrong answer at all. This implies that character-centered representation may cause confusion on solving QA of low difficulty levels.
 
 Figure \ref{fig:qual_result_app_correct} shows an example of our model's prediction about QA of Difficulty 4. It shows the difference between the scores from visual inputs and the scores from scripts. Since there is a word \textit{chat} in the scripts, the model can easily match the context between the scripts and the correct answer. But, in the case of visual inputs, it's hard to find evidence about \texttt{group chat} and has a lower score. Although it has low scores about the correct answer, the final score has the highest score to the correct answer. It means that scores from different modalities complement each other to infer the correct answer.
 
 Figure \ref{fig:qual_result_app_wrong} shows the weakness of character-centered context matching. Only the first candidate has a word \textit{Haeyoung2}, and our model predicts it as a correct answer with high scores. Although it predicts the correct answer in the \texttt{S.High} scores, it cannot infer the correct answer from visual inputs which have the appearance of \textit{Haeyoung2}. Such a question is hard to solve because it requires common sense beyond simply understanding the contents of the video. We can challenge these types of difficulties if a model utilizes object annotations in the background as well as the character-oriented annotations. We expect that our ongoing project for providing hierarchical character-centered story descriptions, objects, and places will be helpful to handle those difficult questions. 
\end{appendix}

\end{document}